%% file: ms.tex
\documentclass[10pt,journal,compsoc]{stylesheet_IEEEtran/IEEEtran}
\ifCLASSOPTIONcompsoc
  \usepackage[nocompress]{cite}
\else
  \usepackage{cite}
\fi
\ifCLASSINFOpdf
\else
\fi
\ifCLASSOPTIONcompsoc
 \usepackage[caption=false,font=footnotesize,labelfont=sf,textfont=sf]{subfig}
\else
 \usepackage[caption=false,font=footnotesize]{subfig}
\fi
\usepackage[english]{babel}
\usepackage[utf8x]{inputenc}
\usepackage[T1]{fontenc}

\usepackage{amsmath}
\usepackage{graphicx}
\usepackage[colorinlistoftodos]{todonotes}
\usepackage[colorlinks=true, allcolors=blue]{hyperref}
\usepackage{xcolor}
\usepackage{amsthm}
\usepackage{caption}
\usepackage{multirow, makecell}
\usepackage{amssymb}
\usepackage{soul}
\usepackage{cleveref}
\usepackage{wrapfig}
\usepackage{tabularx} 
\usepackage[most]{tcolorbox} 
\usepackage{colortbl} 
\usepackage{enumitem} 
\usepackage{booktabs} 
\usepackage{mathabx} 

\definecolor{sankey_gainsboro}{RGB}{227,227,227}
\newtcolorbox[auto counter]{insertbox}[2][0]{
floatplacement=t,float,
arc=0mm,
colback=white,
colframe=white!50!black,
fonttitle=\bfseries,
coltitle=black,
detach title,before upper={\tcbtitle\\[1ex]},
title=Insert~\thetcbcounter: #2,
#1}

\makeatletter
\newcommand\footnoteref[1]{\protected@xdef\@thefnmark{\ref{#1}}\@footnotemark}
\makeatother

\begin{document}
%

\title{Informed Machine Learning --\\A Taxonomy and Survey of Integrating\\Prior Knowledge into Learning Systems}
%
%
%
%

\author{Laura~von~Rueden,
        Sebastian~Mayer,
        Katharina~Beckh,
        Bogdan~Georgiev,
        Sven~Giesselbach,
        Raoul~Heese,
        Birgit~Kirsch,
        Julius~Pfrommer,
        Annika~Pick, 
        Rajkumar~Ramamurthy,
        Michal~Walczak,\\
        Jochen~Garcke,
        Christian~Bauckhage
        and~Jannis~Schuecker
\IEEEcompsocitemizethanks{
\IEEEcompsocthanksitem All authors are with the Fraunhofer Center for Machine Learning.
\IEEEcompsocthanksitem Laura von Rueden, Katharina Beckh, Bogdan Georgiev, Sven Giesselbach, Birgit Kirsch, Annika Pick, Rajkumar Ramamurthy, Christian Bauckhage and Jannis Schuecker are with the Fraunhofer IAIS, Institute for Intelligent Analysis and Information Systems, 53757 Sankt Augustin, Germany.
\IEEEcompsocthanksitem Sebastian Mayer and Jochen Garcke are with the Fraunhofer SCAI, Institute for Algorithms and Scientific Computing, 53757 Sankt Augustin, Germany.%
\IEEEcompsocthanksitem Raoul Heese and Micha{\l} Walczak are with the Fraunhofer ITWM, Institute for Industrial Mathematics, 67663 Kaiserslautern, Germany.%
\IEEEcompsocthanksitem Julius Pfrommer is with the Fraunhofer IOSB, Institute for Optronics, System Technologies and Image Exploitation, 76131 Karlsruhe, Germany.%
\IEEEcompsocthanksitem Corresponding author: laura.von.rueden@iais.fraunhofer.de%
}
}

%
%

\markboth{PREPRINT (ORIGINAL PUBLISHED AT IEEE TRANSACTIONS ON KNOWLEDGE AND DATA ENGINEERING)}{}
%



\IEEEtitleabstractindextext{%
\begin{abstract}
\input{0_abstract.tex}
\end{abstract}

}

\maketitle
\IEEEdisplaynontitleabstractindextext

%
\IEEEpeerreviewmaketitle

\input{2x_concept_figure}
\input{1_introduction.tex} 

\input{2_concept.tex}

\input{4x_taxonomy_figure}
\input{3_classification}

\input{4_taxonomy.tex}

\input{5_approaches.tex}

\input{6_related_work.tex}

\input{7_discussion.tex}

\input{8_conclusion.tex}

\ifCLASSOPTIONcompsoc
 \section*{Acknowledgments}
\else
 \section*{Acknowledgment}
\fi

This work is a joint effort of the Fraunhofer Research Center for Machine Learning (RCML) within the Fraunhofer Cluster of Excellence Cognitive Internet Technologies (CCIT)
and the Competence Center for Machine Learning Rhine Ruhr (ML2R) which is funded by the Federal Ministry of Education and Research of Germany (grant no. 01|S18038B). The team of authors 
gratefully acknowledges this support.

Moreover the authors would like to thank Dorina \mbox{Weichert}, Daniel Paurat, Lars Hillebrand, Theresa Bick and Nico Piatkowski for helpful discussions.



%
%
%

\bibliographystyle{IEEEtran}
\bibliography{informed_ml_clean}

\end{document}

%% file: 0_abstract.tex
Despite its great success, machine learning can have its limits when dealing with insufficient training data. A potential solution is the additional integration of prior knowledge into the training process which leads to the notion of \textit{informed machine learning}.
In this paper, we present a structured overview of various approaches in this field.
We provide a definition and propose a concept for informed machine learning which illustrates its building blocks and distinguishes it from conventional machine learning.
We introduce a taxonomy that serves as a classification framework for informed machine learning approaches. It considers the source of knowledge, its representation, and its integration into the machine learning pipeline.
Based on this taxonomy, we survey related research and describe how different knowledge representations such as algebraic equations, logic rules, or simulation results can be used in learning systems.
This evaluation of numerous papers on the basis of our taxonomy uncovers key methods in the field of informed machine learning.

%% file: 2x_concept_figure.tex
\begin{figure*}
    \centering
    \includegraphics{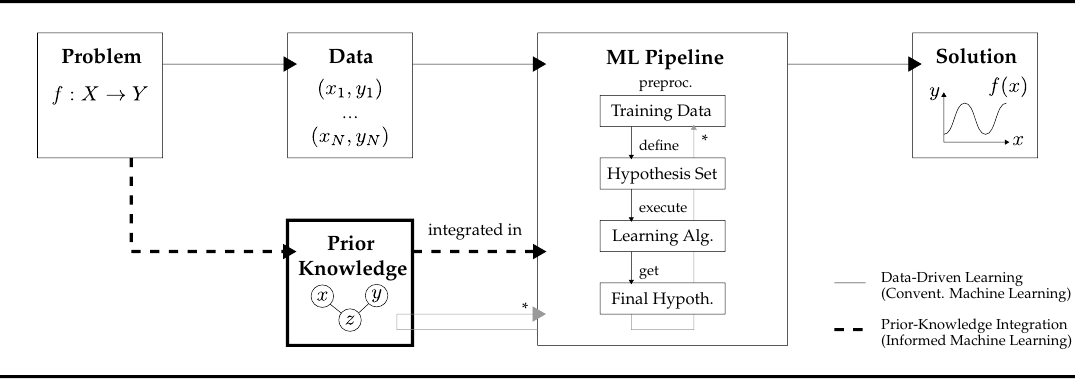}
    \caption{
    \textbf{Information Flow in Informed Machine Learning.}
    The informed machine learning pipeline requires a hybrid information source with two components: Data and prior knowledge.
    In conventional machine learning knowledge is used for data preprocessing and feature engineering, but this process is deeply intertwined with the learning pipeline~(*).
    In contrast, in \textit{informed machine learning}
    prior knowledge comes from an independent source, is given by formal representations (e.g., by knowledge graphs, simulation results, or logic rules), and is explicitly integrated.
    }
    \label{fig:info_flow_iml}
\end{figure*}

%% file: 1_introduction.tex
\IEEEraisesectionheading{
\section{Introduction}
}

\IEEEPARstart{M}{achine} learning has shown great success in building models for pattern recognition in domains ranging from computer vision~\cite{krizhevsky2012imagenet} over speech recognition~\cite{hinton2012deep} and text understanding~\cite{conneau2016very} to Game AI~\cite{silver2016mastering}. In addition to these classical domains, machine learning and in particular deep learning are increasingly important and successful in engineering and the sciences \cite{butler2018machine, ching2018opportunities, kutz2017deep}. These success stories are grounded in the data-based nature of the approach of learning from a tremendous number of examples.

However, there are many circumstances where purely data-driven approaches can reach their limits or lead to unsatisfactory results. 
The most obvious scenario is that not enough data is available to train well-performing and sufficiently generalized models.
Another important aspect is that a purely data-driven model might not meet constraints such as dictated by natural laws, or given through regulatory or security guidelines, which are important for trustworthy~AI~\cite{brundage2020toward}.
With machine learning models becoming more and more complex, there is also a growing need for models to be interpretable and explainable~\cite{roscher2019explainable}.

These issues have led to increased research on how to improve machine learning models by additionally incorporating prior knowledge into the learning process.
Although integrating knowledge into machine learning is common, e.g. through labelling or feature engineering, we observe a growing interest in the integration of more knowledge, and especially of further formal knowledge representations.
For example, logic rules~\cite{diligenti2017integrating,xu2017semantic} or algebraic equations~\cite{karpatne2017physics,stewart2017label} have been added as constraints to loss functions.
Knowledge graphs can enhance neural networks with information about relations between instances~\cite{battaglia2016interaction}, which is of interest in image classification~\cite{marino2017more, jiang2018hybrid}. Furthermore, physical simulations have been used to enrich training data~\cite{cully2015robots,lee2019spigan,pfrommer2018optimisation}.
This heterogeneity in approaches leads to some redundancy in nomenclature; for instance, we find terms such as physics-informed deep learning~\cite{raissi2017physics1}, physics-guided neural networks~\cite{karpatne2017physics}, or semantic-based regularization~\cite{diligenti2017semantic}.
The recent growth of research activities shows that the combination of data- and knowledge-driven approaches becomes relevant in more and more areas.
However, the growing number and increasing variety of research papers in this field motivates a systematic survey.

A recent survey synthesizes this into a new paradigm of theory-guided data science and points out the importance of enforcing scientific consistency in machine learning \cite{karpatne2017theory}.
Even for support vector machines there exists a survey about the incorporation of knowledge into this formalism \cite{lauer2008incorporating}.
The fusion of symbolic and connectionist AI seems more and more approachable.
In this regard, we refer to recent a survey on graph neural networks and a research direction framed as relational inductive bias ~\cite{battaglia2018relational}.
Our work complements the aforementioned surveys by providing a systematic categorization of knowledge representations that are integrated into machine learning.
We provide a structured overview based on a survey of a large number of research papers on how to integrate additional, prior knowledge into the machine learning pipeline.
As an umbrella term for such methods, we henceforth use \textit{informed machine learning}.

Our contributions are threefold:
We propose an abstract concept for informed machine learning that clarifies its building blocks and relation to conventional machine learning. It states that informed learning uses a hybrid information source that consists of data and prior knowledge, which comes from an independent source and is given by formal representations.
Our main contribution is the introduction of a taxonomy that classifies informed machine learning approaches, which is novel and the first of its kind.
It contains the dimensions of the knowledge source, its representation, and its integration into the machine learning pipeline.
We put a special emphasis on categorizing various knowledge representations, since this may enable practitioners to incorporate their domain knowledge into machine learning processes.
Moreover, we present a description of available approaches and explain how different knowledge representations, e.g., algebraic equations, logic rules, or simulation results, can be used in informed machine learning.

Our goal is to equip potential new users of informed machine learning with established and successful methods. As we intend to survey a broad spectrum of methods in this field, we cannot describe all methodical details and we do not claim to have covered all available research papers. We rather aim to analyze and describe common grounds as well as the diversity of approaches in order to identify the main research directions in informed machine learning.

In Section~\ref{sec:concept}, we begin with a formulation of our concept for \textit{informed machine learning}.
In Section~\ref{sec:classification}, we describe how we classified the approaches in terms of our applied surveying methodology and our obtained key insights.
Section~\ref{sec:taxonomy} presents the taxonomy and its elements that we distilled from surveying a large number of research papers.
In Section~\ref{sec:subfields}, we describe the approaches for the integration of knowledge into machine learning classified according to the taxonomy in more detail.
After brief historical account in Section~\ref{sec:background}, we finally discuss future directions in Section~\ref{sec:discussion} and conclude in Section~\ref{sec:conclusion}.

%% file: 2_concept.tex
\section{Concept of Informed Machine Learning}\label{Definition}
\label{sec:concept}

In this section, we present our concept of \textit{informed machine learning}. We first state our notion of knowledge and then present our descriptive definition of its integration into machine learning.

\subsection{Knowledge}

The meaning of knowledge is difficult to define in general and is an ongoing debate in philosophy~\cite{steup2018epistemology, zagzebski2017knowledge, machamer2008blackwell}.
During the generation of knowledge, it first appears as useful information~\cite{fayyad1996data}, which is subsequently validated. People validate information about the world using the brain's inner statistical processing capabilities~\cite{Kahneman2011, lake2017building} or by consulting trusted authorities. Explicit forms of validation are given by empirical studies or scientific experiments~\cite{gauch2003scientific, machamer2008blackwell}.

Here, we assume a computer-scientific perspective and understand knowledge as validated information about relations between entities in certain contexts.
Regarding its use in machine learning, an important aspect of knowledge is its formalization. The degree of formalization depends on whether knowledge has been put into writing, how structured the writing is, and how formal and strict the language is that was used (e.g., natural language vs. mathematical formula). The more formally knowledge is represented, the more easily it can be integrated into machine learning.

\subsection{Integrating Prior Knowledge into Machine Learning}
\label{sec:inf_vs_conv_ml}

Apart from the usual information source in a machine learning pipeline, the training data, one can additionally integrate knowledge.
If this knowledge is pre-existent and independent of learning algorithms, it can be called prior knowledge.
Moreover, such prior knowledge can be given by formal representations, which exist in an external, separated way from the learning problem and the usual training data.
Machine learning that explicitly integrates such knowledge representations will henceforth be called \textit{informed machine learning}.

\theoremstyle{definition}
\newtheorem*{definition}{Definition}
\begin{definition}
\textit{Informed machine learning} describes learning from a hybrid information source that consists of data and prior knowledge.
The prior knowledge comes from an independent source, is given by formal representations, and is explicitly integrated into the machine learning pipeline.
\end{definition}

This notion of informed machine learning thus describes the flow of information in Figure~\ref{fig:info_flow_iml} and is distinct from conventional machine learning.

\subsubsection{Conventional Machine Learning}
\label{sec:general_ml}

Conventional machine learning starts with a specific problem for which there is training data. These are fed into the machine learning pipeline, which delivers a solution.
Problems can typically be formulated as regression tasks where inputs $X$ have to be mapped to outputs $Y$. Training data is generated or collected and then processed by algorithms, which try to approximate the unknown mapping. This pipeline comprises four main components, namely the training data, the hypothesis set, the learning algorithm, and the final hypothesis~\cite{abu2012learning}.

In traditional approaches, knowledge is generally used in the learning pipeline, however, mainly for training data preprocessing (e.g. labelling) or feature engineering. This kind of integration is involved and deeply intertwined with the whole learning pipeline, such as the choice of the hypothesis set or the learning algorithm, as depicted in Figure~\ref{fig:info_flow_iml}. Hence, this knowledge is not really used as an independent source or through separated representations, but is rather used with adaption and as required.

\subsubsection{Informed Machine Learning}

The information flow of informed machine learning comprises an additional prior-knowledge integration and thus consists of two lines originating from the problem, as shown in Figure~\ref{fig:info_flow_iml}. These involve the usual training data and additional prior knowledge. The latter exists independently of the learning task and can be provided in form of logic rules, simulation results, knowledge graphs, etc.

The essence of \textit{informed machine learning} is that this prior knowledge is explicitly integrated into the machine learning pipeline, ideally via clear interfaces defined by the knowledge representations. Theoretically, this applies to each of the four components of the machine learning pipeline.

%% file: 4x_taxonomy_figure.tex
\begin{figure*}
    \centering
    \includegraphics{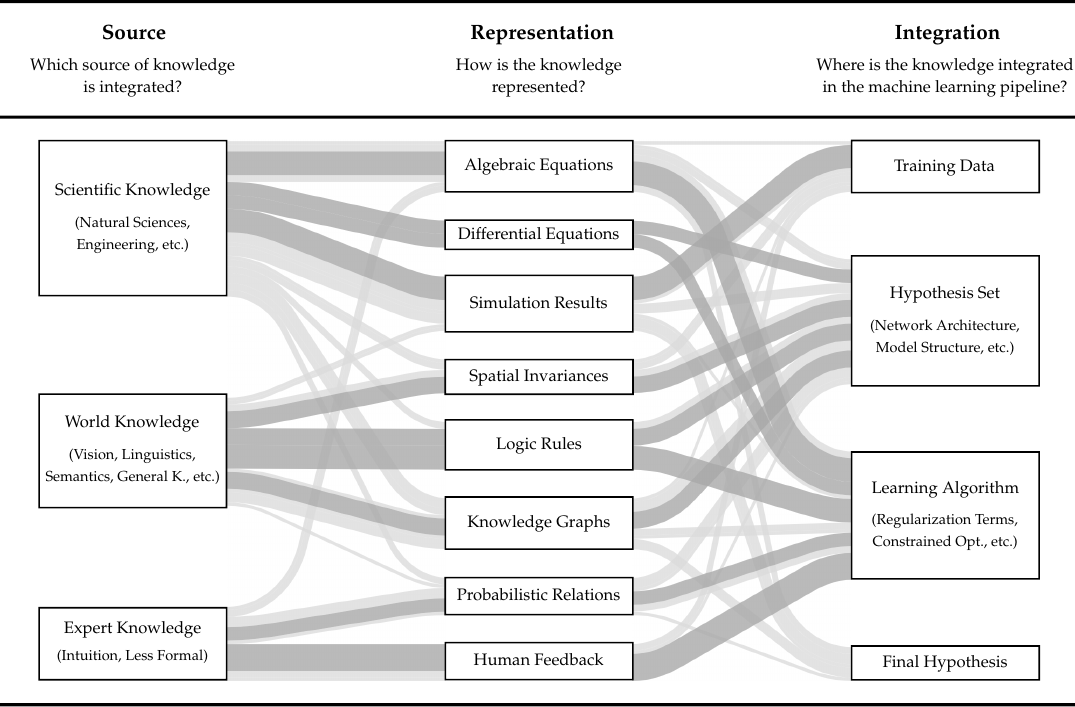}
    \caption{
    \textbf{Taxonomy of Informed Machine Learning.}
    This taxonomy serves as a classification framework for \textit{informed machine learning} and structures approaches according to the three above analysis questions about the \textit{knowledge source}, \textit{knowledge representation} and \textit{knowledge integration}.
    Based on a comparative and iterative literature survey, we identified for each dimension a set of elements that represent a spectrum of different approaches.
    The size of the elements reflects the relative count of papers.
    We combine the taxonomy with a Sankey diagram in which
    the paths connect the elements across the three dimensions and illustrate the approaches that we found in the analyzed papers.
    The broader the path, the more papers we found for that approach.
    Main paths (at least four or more papers with the same approach across all dimensions) are highlighted in darker grey and represent central approaches of informed machine learning.
    }
    \label{fig:taxonomy}
\end{figure*}

%% file: 3_classification.tex
\section{Classification of Approaches}
\label{sec:classification}

To comprehend how the concept of informed machine learning is implemented, we performed a systematic classification of existing approaches based on an extensive literature survey.
Our goals are to uncover different methods, identify their similarities or differences, and to offer guidelines for users and researchers.
In this section, we describe our classification methodology and summarize our key insights.

\subsection{Methodology}

The methodology of our classification is determined by specific analysis questions which we investigated in a systematic literature survey.

\subsubsection{Analysis Questions}

Our guiding question is how prior knowledge can be integrated into the machine learning pipeline.
Our answers will particularly focus on three aspects: Since prior knowledge in informed machine learning consists of an independent source and requires some form of explicit representations, we consider knowledge sources and representations. Since it also is essential at which component of the machine learning pipeline what kind of knowledge is integrated, we also consider integration methods. In short, our literature survey addresses the following three questions: 

\begin{enumerate}
  \setlength\itemsep{0.5em}
    \item \textbf{Source}:\\
    Which source of knowledge is integrated?
    \item \textbf{Representation}:\\
    How is the knowledge represented?
    \item \textbf{Integration}:\\
    Where in the learning pipeline is it integrated?
\end{enumerate}

\subsubsection{Literature Surveying Procedure}

To systematically answer the above analysis questions, we surveyed a large number of publications describing informed machine learning approaches.
We used a comparative and iterative surveying procedure that consisted of different cycles.
In the first cycle, we inspected an initial set of papers and took notes as to how each paper answers our questions. Here, we observed that specific answers occur frequently, which then led to the idea of devising a classification framework in the form of a taxonomy.
In the second cycle, we inspected an extended set of papers and classified them according to a first draft of the taxonomy. We then further refined the taxonomy to match the observations from the literature.
In the third cycle, we re-inspected and re-sorted papers and, furthermore, expanded our set of papers.
This resulted in an extensive literature basis in which all papers are classified according to the distilled taxonomy.

\subsection{Key Insights}
\label{sec:key_insights}

Next, we present an overview over key insights from our systematic classification. As a preview, we refer to Figure~\ref{fig:taxonomy}, which visually summarizes our findings.
A more detailed description of our findings will be given in Sections~\ref{sec:taxonomy} and~\ref{sec:approaches}.

\subsubsection{Taxonomy}
\label{sec:keyinsights_tax}

Based on a comparative and iterative literature survey, we identified a taxonomy that we propose as a classification framework for informed machine learning approaches.
Guided by the above analysis questions, the taxonomy consists of the three dimensions \textit{knowledge source}, \textit{knowledge representation} and \textit{knowledge integration}.
Each dimension contains a set of elements that represent the spectrum of different approaches found in the literature. This is illustrated in the taxonomy in Figure~\ref{fig:taxonomy}.

With respect to  knowledge sources, we found three broad categories: Rather specialized and formalized scientific knowledge, everyday life's world knowledge, and more intuitive expert knowledge. For scientific knowledge we found the most informed machine learning papers. 
With respect to knowledge representations, we found versatile and fine-grained approaches and distilled eight categories (Algebraic equations, differential equations, simulation results, spatial invariances, logic rules, knowledge graphs, probabilistic relations and human feedback).
Regarding knowledge integration, we found approaches for all stages of the machine learning pipeline, from the training data and the hypothesis set, over the learning algorithm, to the final hypothesis. However, most informed machine learning papers consider the two central stages.

Depending on the perspective, the taxonomy can be regarded from either one of two sides: 
An application-oriented user might prefer to read the taxonomy from left to right, starting with some given knowledge source and then selecting representation and integration.
Vice versa, a method-oriented developer or researcher might prefer to read the taxonomy from right to left, starting with some given integration method. 
For both perspectives, knowledge representations are important building blocks and constitute an abstract interface that connects the application- and the method-oriented side.

\subsubsection{Frequent Approaches}

The taxonomy serves as a classification framework and allows us to identify frequent approaches of informed machine learning.
In our literature survey, we categorized each research paper with respect to each of the three taxonomy dimensions.

\textbf{Paths through the Taxonomy.}
When visually highlighting and connecting them, a specific combination of entries across the taxonomy dimensions figuratively results in a path through the taxonomy. Such paths represent specific approaches towards informed learning and we illustrate this by combining the taxonomy with a Sankey diagram, as shown in Figure~\ref{fig:taxonomy}.
We observe that, while various paths through the taxonomy are possible, specific ones occur more frequently and we will call them main paths. 
For example, we often observed the approach that scientific knowledge is represented in algebraic equations, which are then integrated into the learning algorithm, e.g. the loss function. As another example, we often found that world knowledge such as linguistics is represented by logic rules, which are then integrated into the hypothesis set, e.g. the network architecture.
These paths, especially the main paths, can be used as a guideline for users new to the field or provide a set of baseline methods for researchers.

\begin{figure}[t]
    \centering
        \includegraphics[width=\columnwidth]{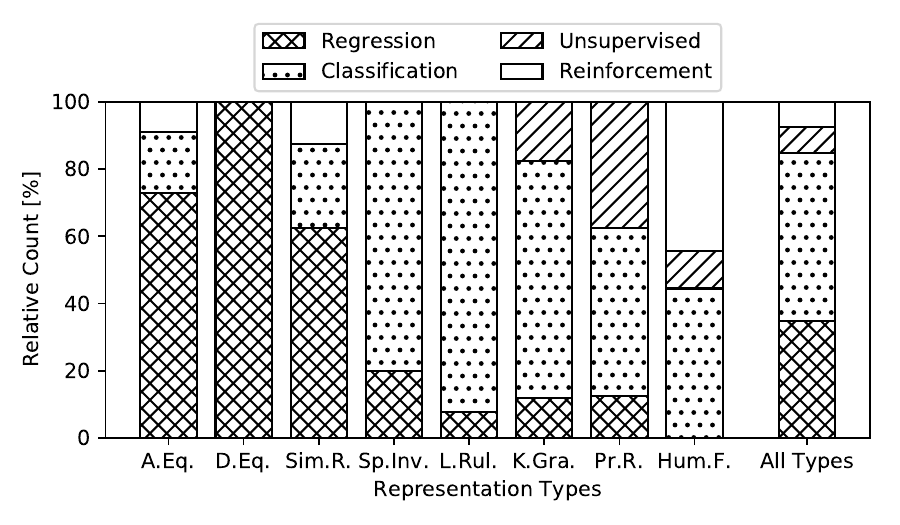}
    \caption{
    \textbf{Knowledge Representations and Learning Tasks.}
    }
    \label{fig:representation_tasks}
\end{figure}

\textbf{Paths from Source to Representation.}
We found that the paths from source to representation form groups. That is, for every knowledge source there appear prevalent representation types.
Scientific knowledge is mainly represented in terms of algebraic or differential equations or exist in form of simulation results. While other forms of representation are possible, too, there is a clear preference for equations or simulations, likely because most sciences aim at finding natural laws encoded in formulas.
For world knowledge, the representation forms of logic rules, knowledge graphs, or spatial invariances are the primary ones. These can be understood as a group of symbolic representations.
Expert knowledge is mainly represented by probabilistic relations or human feedback. This is appears reasonable because such representations allow for informality as well as for a degree of uncertainty, both of which might be useful for representing intuition.
We also performed an additional analysis on the dependency of the learning task and found a confirmation of the above described representation groups as shown in Figure~\ref{fig:representation_tasks}.

From a theoretical point of view, transformations between representations are possible and indeed often apparent within the aforementioned groups. For example, equations can be transformed to simulation results, or logic rules can be represented as knowledge graphs and vice versa. 
Nevertheless, from a practical point of view, differentiating between forms of representations appears useful as specific representations might already be available in a given set up.

\textbf{Paths from Representation to Integration.}
For most of the representation types we found at least one main path to an integration type. The following mappings can be observed. Simulation results are very often integrated into the training data. Knowledge graphs, spatial invariances, and logic rules are frequently incorporated into the hypothesis set. The learning algorithm is mainly enhanced by algebraic or differential equations, logic rules, probabilistic relations, or human feedback. Lastly, the final hypothesis is often checked by knowledge graphs or also by simulation results. However, since we observed various possible types of integration for all representation types, the integration still appears to be problem specific. 

\begin{figure}[t]
    \centering
        \includegraphics[width=\columnwidth]{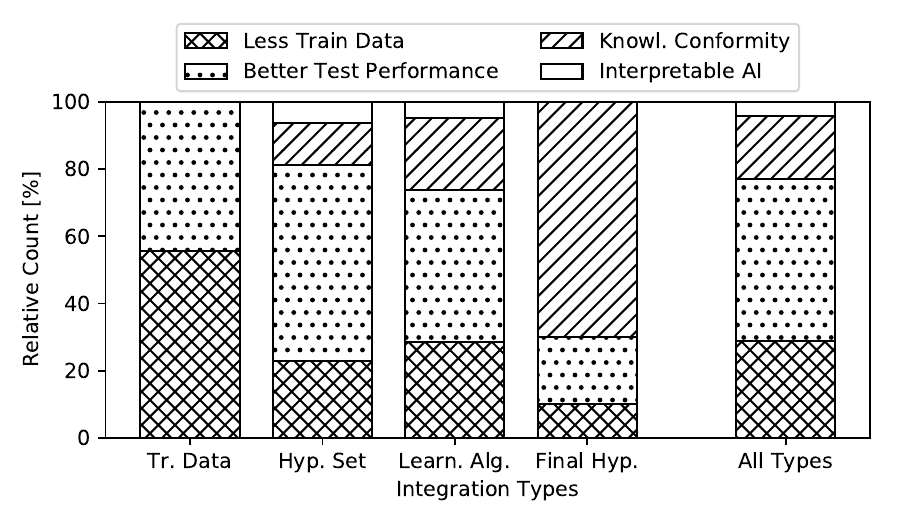}
    \caption{
    \textbf{Knowledge Integration and its Goals.}
    }
    \label{fig:integration_goals}
\end{figure}

Hence, we additionally analyzed the literature for the goal of the prior knowledge integration and found four main goals: Data efficiency, accuracy, interpretability, or knowledge conformity. Although these goals are interrelated or even partially equivalent according to statistical learning theory, it is interesting to examine them as different motivations for the chosen approach. The distribution of goals for the distinct integration types is shown in Figure~\ref{fig:integration_goals}. We observe that the main goal always is to achieve better performance. The integration of prior knowledge into the training data stands out, because its main goal is to train with less data. The integration into the final hypothesis is also special, because it is mainly used to ensure knowledge conformity for secure and trustworthy AI. All in all, this distribution suggests suitable integration approaches depending on the goal.

%% file: 4_taxonomy.tex
\section{Taxonomy}
\label{sec:taxonomy}
\label{sec:taxonmy}

In this section, we describe the \textit{informed machine learning} taxonomy that we distilled as a classification framework in our literature survey. 
For each of the three taxonomy dimensions \textit{knowledge source}, \textit{knowledge representation} and \textit{knowledge integration} we describe the found elements, as shown in Figure~\ref{fig:taxonomy}.
While an extensive approach categorization according to this taxonomy with further concrete examples will be presented in the next section (Section~\ref{sec:approaches}), we here describe the taxonomy on a more conceptual level.
 
\input{4b_knowl_rep_table.tex}

\subsection{Knowledge Source}

The category \textit{knowledge source} refers to the origin of prior knowledge to be integrated in machine learning.
We observe that the source of prior knowledge can be an established knowledge domain but also knowledge from an individual group of people with respective experience.

We find that prior knowledge often stems from the sciences or is a form of world or expert knowledge, as illustrated on the left in Figure~\ref{fig:taxonomy}.
This list is neither complete nor disjoint but intended show a spectrum from more formal to less formal, or explicitly to implicitly validated knowledge.
Although particular knowledge can be assigned to more than one of these sources, the goal of this categorization is to identify paths in our taxonomy that describe frequent approaches of knowledge integration into machine learning. In the following we shortly describe each of the knowledge sources. 

\textbf{Scientific Knowledge.}
We subsume the subjects of science, technology, engineering, and mathematics under \textit{scientific knowledge}. Such knowledge is typically formalized and validated explicitly through scientific experiments. Examples are the universal laws of physics, bio-molecular descriptions of genetic sequences, or material-forming production processes.

\textbf{World Knowledge.}
By \textit{world knowledge} we refer to facts from everyday life that are known to almost everyone and can thus also be called general knowledge. 
It can be more or less formal.
Generally, it can be intuitive and validated implicitly by humans reasoning in the world surrounding them.
Therefore, world knowledge often describes relations of objects or concepts appearing in the world perceived by humans, for instance, the fact that a bird has feathers and can fly.
Moreover, by world knowledge we also subsume linguistics. Such knowledge can also be explicitly validated through empirical studies. Examples are the syntax and semantics of language.

\textbf{Expert Knowledge.}
We consider \textit{expert knowledge} to be knowledge that is held by a particular group of experts. Within the expert's community it can also be called common knowledge. Such knowledge is rather informal and needs to be formalized, e.g., with human-machine interfaces.
It is also validated implicitly through a group of experienced specialists. In the context of cognitive science, this expert knowledge can also become intuitive~\cite{Kahneman2011}.
For example, an engineer or a physician acquires knowledge over several years of experience working in a specific field. 

\subsection{Knowledge Representation}

The category \textit{knowledge representation} describes how knowledge is formally represented. With respect to the flow of information in informed machine learning in Figure~\ref{fig:info_flow_iml}, it directly corresponds to our key element of prior knowledge. This category constitutes the central building block of our taxonomy, because it determines the potential interface to the machine learning pipeline.

In our literature survey, we frequently encountered certain representation types, as listed in the taxonomy in Figure~\ref{fig:taxonomy} and illustrated more concretely in Table~\ref{tab:rep}.
Our goal is to provide a classification framework of informed machine learning approaches including the used knowledge representation types. Although some types can be mathematically transformed into each other, we keep the representation that are closest to those in the reviewed literature.
Here we give a first conceptual overview over these types.

\textbf{Algebraic Equations.}
Algebraic equations represent knowledge as equality or inequality relations between mathematical expressions consisting of variables or constants. 
Equations can be used to describe general functions or to constrain variables to a feasible set and are thus sometimes also called algebraic constraints.
Prominent examples in Table~\ref{tab:rep} are the equation for the mass-energy equivalence and the inequality stating that nothing can travel faster than the speed of light in vacuum.

\textbf{Differential Equations.}
Differential equations are a subset of algebraic equations, which describe relations between functions and their spatial or temporal derivatives. Two famous examples in Table~\ref{tab:rep} are the heat equation, which is a partial differential equation (PDE), and Newton's second law, which is an ordinary differential equation (ODE).
In both cases, there exists a (possibly empty) set of functions that solve the differential equation for given initial or boundary conditions.
Differential equations are often the basis of a numerical computer simulation. We distinguish the taxonomy categories of differential equations and simulation results in the sense that the former represents a compact mathematical model while the latter represents unfolded, data-based computation results.

\textbf{Simulation Results.}
Simulation results describe the numerical outcome of a computer simulation, which is an approximate imitation of the behavior of a real-world process.
A simulation engine typically solves a mathematical model using numerical methods and produces results for situation-specific parameters.
Its numerical outcome is the simulation result that we describe here as the final knowledge representation.
Examples are the flow field of a simulated fluid or pictures of simulated traffic scenes.

\textbf{Spatial Invariances.}
Spatial invariances describe properties that do not change under mathematical transformations such as translations and rotations. If a geometric object is invariant under such transformations, it has a symmetry (for example, a rotationally symmetric triangle). A function can be called invariant, if it has the same result for a symmetric transformation of its argument.
Connected to invariance is the property of equivariance.

\textbf{Logic Rules.}
Logic provides a way of formalizing knowledge about facts and dependencies and allows for translating ordinary language statements (e.g., $\texttt{IF}\ A\ \texttt{THEN}\ B$) into formal logic rules ($A \Rightarrow B$). Generally, a logic rule consists of a set of Boolean expressions ($A$, $B$) combined with logical connectives ($\land$, $\lor$, $\Rightarrow$, $\dots$).
Logic rules can be also called logic constraints or logic sentences.

\textbf{Knowledge Graphs.}
A graph is a pair $(V,E)$, where $V$ are its vertices and $E$ denotes edges. In a knowledge graph, vertices (or nodes) usually describe concepts whereas edges represent (abstract) relations between them (as in the example ``Man wears shirt'' in Table~\ref{tab:rep}).
In an ordinary weighted graph, edges quantify the strength and the sign of a relationship between nodes.

\textbf{Probabilistic Relations.}
The core concept of probabilistic relations is a random variable $X$ from which samples $x$ can be drawn according to an underlying probability distribution $P(X)$. Two or more random variables $X, Y$ can be interdependent with joint distribution $(x,y) \sim P(X,Y)$.
Prior knowledge could be assumptions on the conditional independence or the correlation structure of random variables or even a full description of the joint probability distributions.

\textbf{Human Feedback.}
Human feedback refers to technologies that transform knowledge via direct interfaces between users and machines.
The choice of input modalities determines the way information is transmitted. Typical modalities include keyboard, mouse, and touchscreen, followed by speech and computer vision, e.g., tracking devices for motion capturing.
In theory, knowledge can also be transferred directly via brain signals using brain-computer interfaces.

\subsection{Knowledge Integration}

The category \textit{knowledge integration} describes where the knowledge is integrated into the machine learning pipeline.

Our literature survey revealed that integration approaches can be structured according to the four components of training data, hypothesis set, learning algorithm, and final hypothesis. Though we present these approaches more thoroughly in Section~\ref{sec:subfields}, the following gives a first conceptual overview. 

\textbf{Training Data.}
A standard way of incorporating knowledge into machine learning is to embody it in the underlying training data.
Whereas a classic approach in traditional machine learning is feature engineering where appropriate features are created from expertise, an informed approach according to our definition is the use of hybrid information in terms of the original data set and an additional, separate source of prior knowledge. This separate source of prior knowledge allows to accumulate information and therefore can create a second data set, which can then be used together with, or in addition to, the original training data.
A prominent approach is simulation-assisted machine learning where the training data is augmented through simulation results.

\textbf{Hypothesis Set.}
Integrating knowledge into the hypothesis set is common, say, through the definition of a neural network's architecture and hyper-parameters. For example, a convolutional neural network applies knowledge as to location and translation invariance of objects in images.
More generally, knowledge can be integrated by choosing model structure. 
A notable example is the design of a network architecture considering a mapping of knowledge elements, such as symbols of a logic rule, to particular neurons.

\textbf{Learning Algorithm.}
Learning algorithms typically involve a loss function that can be modified according to additional knowledge, e.g.~by designing an appropriate regularizer.
A typical approach of informed machine learning is that prior knowledge in form of algebraic equations, for example laws of physics, is integrated by means of additional loss terms.

\textbf{Final Hypothesis.}
The output of a learning pipeline, i.e. the final hypothesis, can be benchmarked or validated against existing knowledge.
For example, predictions that do not agree with known constraints can be discarded or marked as suspicious so that results are consistent with prior knowledge.

%% file: 4b_knowl_rep_table.tex

\newcolumntype{Y}{>{\centering\arraybackslash}X}
\renewcommand\tabularxcolumn[1]{m{#1}}
\begin{table*}[!ht]
\renewcommand{\arraystretch}{1.3}
\label{table_example}
\caption{\textbf{Illustrative Overview of Knowledge Representations in the Informed Machine Learning Taxonomy.}
Each representation type is illustrated by a simple or prominent example
in order to give a first intuitive understanding.
}
    \begin{tabularx}{\textwidth}{YYYYYYYY}
    \toprule
    \bfseries Algebraic&
    \bfseries Differential&
    \bfseries Simulation&
    \bfseries Spatial&
    \bfseries Logic&
    \bfseries Knowledge&
    \bfseries Probabilistic&
    \bfseries Human
    \\
    \bfseries Equations&
    \bfseries Equations&
    \bfseries Results&
    \bfseries Invariances&
    \bfseries Rules&
    \bfseries Graphs&
    \bfseries Relations&
    \bfseries Feedback \\
    \midrule
    {
    \renewcommand{\arraystretch}{2}
    \begin{tabular}{@{}c@{}}
    $E = m \cdot c^2$\\
    $v \leq c$
    \end{tabular}
    }
    &
    {
    \renewcommand{\arraystretch}{2}
    \begin{tabular}{@{}c@{}}
    $\frac{\partial u}{\partial t} = \alpha \frac{\partial^2 u}{\partial x^2}$\\
    $F(x) = m \frac{d^2 x}{d t^2}$
    \end{tabular}
    }
    &
    \includegraphics{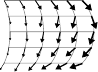}
    &
    \includegraphics{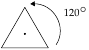}
    &
    $A \land B \Rightarrow C$
    &
    \includegraphics{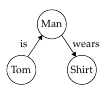}
    &
    \includegraphics{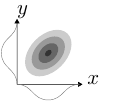}
    &
    \includegraphics{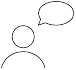}
    \\
    \bottomrule
    \end{tabularx}
\label{tab:rep}
\end{table*}

%% file: 5_approaches.tex
\section{Description of Integration Approaches}
\label{sec:subfields}
\label{sec:approaches}

In this section, we give a detailed account of the informed machine learning approaches we found in our literature survey. We will focus on methods and therefore structure our presentation according to knowledge representations.
This is motivated by the assumption that similar representations are integrated into machine learning in similar ways as they form the mathematical basis for the integration.
Moreover the representations combine both the application- and the method-oriented perspective as described in Section~\ref{sec:keyinsights_tax}.

For each knowledge representation, we describe the informed machine learning approaches in a separate subsection and present the observed (paths from) knowledge source and the observed (paths to) knowledge integration. We describe each dimension along its entities starting with the main path entity, i.e. the one we found in most papers. 

This whole section refers to Table~~\ref{tab:src2rep} and~\ref{tab:rep2int}, which lists the paper references sorted according to our taxonomy.

\input{5a_algebraic_equations.tex}
\input{5b_differential_equations.tex}
\input{5x_table_commands}
\input{5x_new_tables}

\input{5c_simulations.tex}
\input{5d_local_invariances.tex}
\input{5e_logic_rules.tex}
\input{5f_knowledge_graphs.tex}
\input{5g_probabilisitic_relations.tex}
\input{5h_human_feedback.tex}

%% file: 5a_algebraic_equations.tex
\subsection{Algebraic Equations}
\label{sec:Algebraic equations}
\label{sec:rep_alg_eqn}

The main path for algebraic equations that we found in our literature survey comes from scientific knowledge and goes into the learning algorithm, but also other integration types are possible, as illustrated in the following figure.

\begin{figure}[!h]
    \centering
    \includegraphics{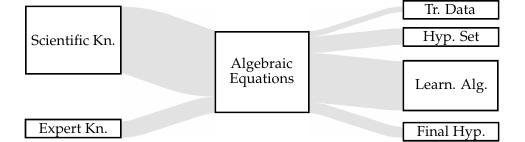}
\end{figure}

\subsubsection{(Paths from) Knowledge Source}

Algebraic equations are mainly used to represent formalized scientific knowledge, but may also be used to express more intuitive expert knowledge. 

\textbf{Scientific Knowledge.}
We observed that algebraic equations are used in machine learning in various domains of natural sciences and engineering, particularly in physics~\cite{karpatne2017physics, muralidhar2018incorporating, lu2017physics, stewart2017label, heese2019good}, but also in biology~\cite{fung2003knowledge, mangasarian2008nonlinear}, robotics~\cite{ramamurthy2019leveraging}, or manufacturing and production processes~\cite{vonkurnatowski2020compensating,lu2017physics}.

Three representative examples are the following:
The trajectory of objects can be described with kinematic laws, e.g., that the position~$y$ of a falling object can be described as a function of time $t$, namely $y(t) = y_0 + v_0 t + at^2$.

Such knowledge from Newtonian mechanics can be used to improve object detection and tracking in videos~\cite{stewart2017label}.
Or, the proportionality of two variables can be expressed via inequality constraints, for example, that the water density $\rho$ at two different depths $d_1 < d_2$ in a lake must obey $\rho(d_1) \leq \rho(d_2)$,
which can be used in water temperature prediction~\cite{karpatne2017physics}.
Furthermore, for the prediction of key performance indicators in production processes, relations between control parameters (e.g. voltage, pulse duration) and intermediate observables (e.g. current density) are known to influence outcomes and can be expressed as linear equations derived from principles of physical chemistry~\cite{lu2017physics}.

\textbf{Expert Knowledge.}
An example for the representation of expert knowledge is to define valid ranges of variables according to experts' intuition as approximation constraints~\cite{muralidhar2018incorporating} or monotonicity constraints~\cite{vonkurnatowski2020compensating}.

\subsubsection{(Paths to) Knowledge Integration}

We observe that a frequent way of integrating equation-based knowledge into machine learning is via the learning algorithm.
The integration into the other stages is possible, too, and we describe the approaches here ordered by their occurence.

\textbf{Learning Algorithm.}
Algebraic equations and inequations can be integrated into learning algorithms via additional loss terms~\cite{stewart2017label,karpatne2017physics,muralidhar2018incorporating, heese2019good}
or, more generally, via constrained problem formulation~\cite{fung2003knowledge, mangasarian2008nonlinear, vonkurnatowski2020compensating}.

\begin{insertbox}[label={in:loss},floatplacement=t]{Knowledge-Based Loss Term}
When learning a function $f^*$ from data $(x_i, y_i)$ where the $x_i$ are input features and the $y_i$ are labels, a knowledge-based loss term $L_k$ can be built into the objective function~\cite{karpatne2017physics, diligenti2017integrating}:
\begin{align}
\begin{split}
f^* = \arg\min_{f} \Big(
& \overbrace{ \lambda_l \textstyle\sum_i{ L(f(x_i),y_i)} }^\texttt{Label-based}
+ \overbrace{ \lambda_r R(f) }^\texttt{Regul.}\\
+ & \underbrace{\lambda_{k} L_k(f(x_i),x_i) }_{\texttt{Knowledge-based}}
\Big)
\end{split}
\label{eq:lossfunction}
\end{align}
Whereas $L$ is the usual label-based loss and $R$ is a regularization function, $L_k$ quantifies the violation of given prior-knowledge equations. Parameters $\lambda_l$, $\lambda_r$ and $\lambda_{k}$ determine the weight of the terms.

Note that $L_k$ only depends on the input features $x_i$ and the learned function $f$ and thus offers the possibility of label-free supervision~\cite{stewart2017label}.
\end{insertbox}

The integration of algebraic equations as knowledge-based loss terms into the learning objective function is detailed in Insert~\ref{in:loss}. These knowledge-based terms measure potential inconsistencies w.r.t., say, physical laws~\cite{karpatne2017physics, stewart2017label}.
Such an extended loss is usually called physics-based or hybrid loss and fosters the learning from data as well as from prior knowledge.
Beyond the measuring inconsistencies with exact formulas, inconsistencies with approximation ranges or general monotonicity constraints, too, can be quantified via rectified linear units~\cite{muralidhar2018incorporating}.

As a further approach, support vector machines can incorporate knowledge by relaxing the optimization problem into a linear minimization problem to which constraints are added in form of linear inequalities~\cite{fung2003knowledge}.
Similarly, it is possible to relax the optimization problem behind certain kernel-based approximation methods to constrain the behavior of a regressor or classifier in a possibly nonlinear region of the input domain~\cite{mangasarian2008nonlinear}.

\textbf{Hypothesis Set.}
An alternative approach is the integration into the hypothesis set. In particular, algebraic equations can be translated into the architecture of neural networks~\cite{bauckhage2018informed, lu2017physics, ramamurthy2019leveraging}.
One idea is to sequence predefined operations leading to a functional decomposition~\cite{bauckhage2018informed}.
More specifically, relations between input parameters, intermediate observables, or output variables reflecting physical constraints can be encoded as linear connections between the layers of a network model ~\cite{lu2017physics, ramamurthy2019leveraging}.

\textbf{Final Hypothesis.}
Another integration path applies algebraic equations to the final hypothesis, mainly serving as a consistency check with given constraints from a knowledge domain. This can be implemented as an inconsistency measure that quantifies the deviation of the predicted results from given knowledge similar to the above knowledge-based loss terms. It can then be used as an additional performance metric for model comparison~\cite{karpatne2017physics}. Such a physical consistency check can also comprise an entire diagnostics set of functions describing particular characteristics~\cite{king2018deep}.

\textbf{Training Data.}
Another natural way of integrating algebraic equations into machine learning is to use them for training data generation. While there are many papers in this category, we want to highlight one that integrates prior knowledge as an independent, second source of information by constructing a specific feature vector that directly models physical properties and constraints~\cite{jeong2015data}.

%% file: 5b_differential_equations.tex
\subsection{Differential Equations}
\label{sec:Differential Equations}
\label{sec:rep_dif_eqn}

Next, we describe informed machine learning approaches based on differential equations,
which frequently represent scientific knowledge and are integrated into the hypothesis set or the learning algorithm.

\begin{figure}[!h]
    \centering
    \includegraphics{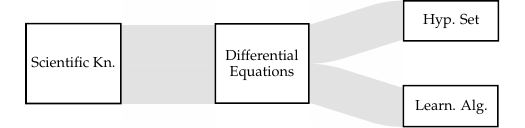}
\end{figure} 

\subsubsection{(Paths from) Knowledge Source}

Differential equations model the behavior of dynamical systems by relating state variables to their rate of change. In the literature discussed here, differential equations represent knowledge from the natural sciences.

\textbf{Scientific Knowledge.}
Here we give three prominent examples:
The work in \cite{raissi2017physics1,yang2018physics} considers the Burger's equation, which is used in fluid dynamics to model simple one-dimensional currents and in traffic engineering to describe traffic density behavior.
Advection-diffusion equations~\cite{debezenac} are used in oceanography to model the evolution of sea surface temperatures.
The Schr\"odinger equation studied in \cite{raissi2017physics1} describes quantum mechanical phenomena such as wave propagation in optical fibres or the behavior of Bose-Einstein condensates.

\subsubsection{(Paths to) Knowledge Integration}

Regarding the integration of differential equations, our survey particularly focuses on the integration into neural network models.

\textbf{Learning Algorithm. }
A neural network can be trained to approximate the solution of a differential equation. To this end, the governing differential equation is integrated into the loss function similar to Equation~\ref{eq:lossfunction}~\cite{lagaris1998artificial}. This requires evaluating derivatives of the network with respect to its inputs, for example, via automatic differentiation, an approach that was recently adapted to deep learning \cite{raissi2017physics1}. This ensures the physical plausibility of the neural network output. An extension to generative models is possible, too \cite{yang2018physics}. Finally, probabilistic models can also be trained by minimizing the distance between the model conditional density and the Boltzmann distribution dictated by a differential equation and boundary conditions \cite{zhu2019}.

\textbf{Hypothesis Set.}
In many applications, differential equations contain unknown time- and space-dependent parameters. Neural networks can model the behavior of such parameters, which then leads to hybrid architectures where the functional form of certain components is analytically derived from (partially) solving differential equations~\cite{psichogios1992hybrid, lutter2019deep,debezenac}. In other applications, one faces the problem of unknown mappings from input data to quantities whose dynamics are governed by known differential equations, usually called system states. Here, neural networks can learn a mapping from observed data to system states~\cite{belbute-peres2018}. This also leads to hybrid architectures with knowledge-based modules, e.g. in form of a physics engine.

%% file: 5x_table_commands.tex
\newboolean{show_tentative}
\setboolean{show_tentative}{true}
\newcommand\tenta[1]{\ifthenelse{\boolean{show_tentative}}{\textcolor{black}{#1}}{\hide{#1}}}



\newcommand\nat{\cite{jeong2015data}}
\newcommand\ndt{}
\newcommand\nst{\cite{lerer2016learning, shrivastava2017learning, lee2019spigan, pfrommer2018optimisation, karpatne2017physics, deist2019simulation, rai2019using}}
\newcommand\nit{\cite{wu2018physics}}
\newcommand\nlt{}
\newcommand\nkt{}
\newcommand\npt{}
\newcommand\nht{}


\newcommand\nah{\cite{ramamurthy2019leveraging, lu2017physics}}
\newcommand\ndh{\cite{debezenac, lutter2019deep},\tenta{\cite{psichogios1992hybrid,belbute-peres2018}}}
\newcommand\nsh{\cite{wang1997knowledge, leary2003knowledge},\tenta{\cite{kim2007hybrid}}}
\newcommand\nih{\cite{ling2016reynolds, butter2018deep}}
\newcommand\nlh{\cite{towell1994knowledge, garcez1999connectionist}}
\newcommand\nkh{\cite{battaglia2016interaction, chang2016compositional, choi2017gram}}
\newcommand\nph{}
\newcommand\nhh{}


\newcommand\nal{\cite{karpatne2017physics, stewart2017label, muralidhar2018incorporating, heese2019good, fung2003knowledge, mangasarian2008nonlinear}}
\newcommand\ndl{\cite{lagaris1998artificial,raissi2017physics1,yang2018physics},\tenta{\cite{zhu2019}}}
\newcommand\nsl{\cite{du2018learning}}
\newcommand\nil{}
\newcommand\nll{}
\newcommand\nkl{\cite{ma2018multi,che2015deep}}
\newcommand\npl{\cite{messaoud2009integrating, borboudakis2012incorporating}}
\newcommand\nhl{}


\newcommand\naf{\cite{karpatne2017physics, king2018deep}}
\newcommand\ndf{}
\newcommand\nsf{\cite{hautier2010finding, du2018learning, pfrommer2018optimisation}}
\newcommand\nif{}
\newcommand\nlf{}
\newcommand\nkf{}
\newcommand\npf{}
\newcommand\nhf{}



\newcommand\sat{}
\newcommand\sdt{}
\newcommand\sst{}
\newcommand\sit{}
\newcommand\slt{}
\newcommand\skt{}
\newcommand\spt{}
\newcommand\sht{}


\newcommand\sah{}
\newcommand\sdh{}
\newcommand\ssh{}
\newcommand\sih{}
\newcommand\slh{\cite{richardson2006markov,kimmig2012short}}
\newcommand\skh{}
\newcommand\sph{}
\newcommand\shh{}


\newcommand\sal{}
\newcommand\sdl{}
\newcommand\ssl{}
\newcommand\sil{}
\newcommand\sll{\cite{chang2007guiding, hu2016deep, hu2016harnessing}}
\newcommand\skl{}
\newcommand\spl{}
\newcommand\shl{}


\newcommand\saf{}
\newcommand\sdf{}
\newcommand\ssf{}
\newcommand\sif{}
\newcommand\slf{}
\newcommand\skf{\tenta{\cite{glavavs2018explicit, mrkvsic2016counter}}}
\newcommand\spf{}
\newcommand\shf{}



\newcommand\wat{}
\newcommand\wdt{}
\newcommand\wst{\cite{shrivastava2017learning}}
\newcommand\wit{\cite{niyogi1998incorporating, bergman2019symmetry}}
\newcommand\wlt{}
\newcommand\wkt{\cite{mintz2009distant}}
\newcommand\wpt{}
\newcommand\wht{}


\newcommand\wah{}
\newcommand\wdh{}
\newcommand\wsh{}
\newcommand\wih{\cite{cohen2016group, dieleman2016exploiting, li2018deep, worrall2017harmonic, scholkopf1998prior}}
\newcommand\wlh{\cite{francca2014fast,schiegg2012markov, sachan2018learning}}
\newcommand\wkh{\cite{jiang2018hybrid, marino2017more, liang2018symbolic,zhou2018commonsense}}
\newcommand\wph{\cite{yet2014combining}}
\newcommand\whh{}


\newcommand\wal{}
\newcommand\wdf{}
\newcommand\wsl{}
\newcommand\wil{
}
\newcommand\wll{\cite{diligenti2017semantic, diligenti2017integrating, stewart2017label, xu2017semantic}}
\newcommand\wkl{\cite{che2015deep},\tenta{\cite{bian2014knowledge,zhang2019ernie}}}
\newcommand\wpl{}
\newcommand\whf{}


\newcommand\waf{}
\renewcommand\wdf{}
\newcommand\wsf{}
\newcommand\wif{}
\newcommand\wlf{}
\newcommand\wkf{\cite{fang2017object, vonrueden2020towards}}
\newcommand\wpf{}



\newcommand\eat{}
\newcommand\edt{}
\newcommand\est{}
\newcommand\eit{}
\newcommand\elt{}
\newcommand\ekt{}
\newcommand\ept{}
\newcommand\eht{\cite{hester2018deep, kaplan2017beating}}


\newcommand\eah{\cite{bauckhage2018informed}}
\newcommand\edh{}
\newcommand\esh{}
\newcommand\eih{}
\newcommand\elh{}
\newcommand\ekh{}
\newcommand\eph{\cite{constantinou2016integrating, heckerman1995learning,yet2014combining}}
\newcommand\ehh{\cite{hester2018deep}}


\newcommand\eal{\cite{muralidhar2018incorporating}}
\newcommand\edl{}
\newcommand\esl{}
\newcommand\eil{}
\renewcommand\ell{}
\newcommand\ekl{}
\newcommand\epl{\cite{feelders2006learning,Campos2007bayesian,heckerman1995learning,richardson2003learning}}
\newcommand\ehl{\cite{brown2012dis, fails2003interactive, choo2013utopian, rieger2019interpretations, schramowski2020right, knox2009interactively, christiano2017deep, kaplan2017beating}}


\newcommand\eaf{}
\newcommand\edf{}
\newcommand\esf{}
\newcommand\eif{}
\newcommand\elf{}
\newcommand\ekf{}
\newcommand\epf{\cite{yet2014not}}
\newcommand\ehf{}

%% file: 5x_new_tables.tex
\newcolumntype{Y}{>{\raggedright\arraybackslash}X}


\begin{table*}
\caption{
\textbf{
References Classified by Knowledge Representation and (Path from) Knowledge Source.
}
}
\renewcommand{\arraystretch}{1.3}
\scriptsize
\centering
    \begin{tabularx}{\textwidth}{@{}lllllllll@{}}
    \toprule
    \textbf{SOURCE}
    &
    \multicolumn{8}{c}{\textbf{REPRESENTATION}} \\
    \cmidrule{2-9}
    &
    Algebraic&
    Differential&
    Simulation&
    Spatial&
    Logic&
    Knowledge&
    Probabilistic&
    Human
    \\
    &
    Equations&
    Equations&
    Results&
    Invariances&
    Rules&
    Graphs&
    Relations&
    Feedback
    \\
    \midrule
    Scientific
    &
    \cite{stewart2017label, karpatne2017physics, muralidhar2018incorporating}
    &
    \cite{raissi2017physics1, yang2018physics, debezenac}
    &
    \cite{karpatne2017physics, lee2019spigan, pfrommer2018optimisation}
    &
    \cite{ling2016reynolds, butter2018deep,wu2018physics}
    &
    \cite{towell1994knowledge, garcez1999connectionist}
    &
    \cite{battaglia2016interaction, ma2018multi,che2015deep}
    &
    \cite{messaoud2009integrating, borboudakis2012incorporating}
    &
    \\
    Knowledge
    &
    \cite{lu2017physics, heese2019good, fung2003knowledge}
    &
    \cite{lagaris1998artificial,zhu2019, psichogios1992hybrid}
    &
    \cite{deist2019simulation,kim2007hybrid, hautier2010finding}
    &
    &
    &
    \cite{chang2016compositional, choi2017gram}
    &
    &
    \\
    &
    \cite{ramamurthy2019leveraging, mangasarian2008nonlinear, king2018deep}
    &
    \cite{belbute-peres2018, lutter2019deep}
    &
    \cite{lerer2016learning, rai2019using, du2018learning}
    &
    &
    &
    &
    &
    \\
    &
    \cite{jeong2015data, vonkurnatowski2020compensating}
    &
    &
    \cite{shrivastava2017learning, wang1997knowledge, leary2003knowledge}
    &
    &
    &
    &
    &
    \\
    \midrule
    World
    &
    &
    &
    \cite{shrivastava2017learning}
    &
    \cite{cohen2016group, dieleman2016exploiting, scholkopf1998prior}
    &
    \cite{diligenti2017integrating, stewart2017label, xu2017semantic}
    &
    \cite{marino2017more, jiang2018hybrid,che2015deep}
    &
    \cite{yet2014combining}
    &
    \\
    Knowledge
    &
    &
    &
    &
    \cite{li2018deep, worrall2017harmonic, niyogi1998incorporating}
    &
    \cite{diligenti2017semantic, schiegg2012markov, sachan2018learning}
    &
    \cite{zhou2018commonsense,mintz2009distant,liang2018symbolic}
    &
    &
    \\
    &
    &
    &
    &
    \cite{bergman2019symmetry}
    &
    \cite{chang2007guiding, hu2016deep, hu2016harnessing}
    &
    \cite{zhang2019ernie,mrkvsic2016counter,bian2014knowledge}
    &
    &
    \\
    &
    &
    &
    &
    &
    \cite{francca2014fast, richardson2006markov,kimmig2012short}
    &
    \cite{glavavs2018explicit, fang2017object,peters2019knowledge}
    &
    &
    \\
    \midrule
    Expert
    &
    \cite{bauckhage2018informed, muralidhar2018incorporating, vonkurnatowski2020compensating}
    &
    &
    &
    &
    &
    &
    \cite{Campos2007bayesian,constantinou2016integrating,yet2014combining}
    &
    \cite{choo2013utopian,knox2009interactively, kaplan2017beating}
    \\
    Knowledge
    &
    &
    &
    &
    &
    &
    &
    \cite{heckerman1995learning,richardson2003learning,feelders2006learning}
    &
    \cite{christiano2017deep,hester2018deep,brown2012dis}
    \\
    &
    &
    &
    &
    &
    &
    &
    \cite{yet2014not}
    &
    \cite{fails2003interactive,  rieger2019interpretations, schramowski2020right}
    \\
    &
    &
    &
    &
    &
    &
    &
    &
    \\
    \bottomrule
    \\
    \end{tabularx}
\label{tab:src2rep}
\end{table*}



\begin{table*}
\caption{
\textbf{
References Classified by Knowledge Representation and (Path to) Knowledge Integration.
}
}
\renewcommand{\arraystretch}{1.3}
\scriptsize
\centering
    \begin{tabularx}{\textwidth}{@{}lllllllll@{}}
    \toprule
    \textbf{INTEGRAT.}
    &
    \multicolumn{8}{c}{\textbf{REPRESENTATION}} \\
    \cmidrule{2-9}
    &
    Algebraic&
    Differential&
    Simulation&
    Spatial&
    Logic&
    Knowledge&
    Probabilistic&
    Human
    \\
    &
    Equations&
    Equations&
    Results&
    Invariances&
    Rules&
    Graphs&
    Relations&
    Feedback
    \\
    \midrule
    Training
    &
    \cite{jeong2015data}
    &
    &
    \cite{karpatne2017physics,lee2019spigan, pfrommer2018optimisation}
    &
    \cite{wu2018physics, niyogi1998incorporating, bergman2019symmetry}
    &
    &
    \cite{mintz2009distant}
    &
    &
    \cite{hester2018deep, kaplan2017beating}
    \\
    Data
    &
    &
    &
    \cite{deist2019simulation, lerer2016learning, rai2019using}
    &
    &
    &
    &
    &
    \\
    &
    &
    &
    \cite{shrivastava2017learning}
    &
    &
    &
    &
    &
    \\
    \midrule
    Hypothesis
    &
    \cite{ramamurthy2019leveraging, lu2017physics, bauckhage2018informed}
    &
    \cite{debezenac, lutter2019deep, psichogios1992hybrid}
    &
    \cite{wang1997knowledge, leary2003knowledge, kim2007hybrid}
    &
    \cite{scholkopf1998prior,ling2016reynolds, butter2018deep}
    &
    \cite{towell1994knowledge,schiegg2012markov, sachan2018learning}
    &
    \cite{battaglia2016interaction,marino2017more, jiang2018hybrid}
    &
    \cite{constantinou2016integrating, heckerman1995learning,yet2014combining}
    &
    \cite{hester2018deep}
    \\
    Set
    &
    &
    \cite{belbute-peres2018}
    &
    &
    \cite{cohen2016group, dieleman2016exploiting, li2018deep}
    &
    \cite{kimmig2012short,richardson2006markov,garcez1999connectionist}
    &
    \cite{choi2017gram, chang2016compositional, liang2018symbolic}
    &
    &
    \\
    &
    &
    &
    &
    \cite{worrall2017harmonic}
    &
    \cite{francca2014fast}
    &
    \cite{zhou2018commonsense,zhang2019ernie, peters2019knowledge}
    &
    &
    \\
    \midrule
    Learning
    &
    \cite{karpatne2017physics, stewart2017label, muralidhar2018incorporating}
    &
    \cite{lagaris1998artificial,raissi2017physics1,yang2018physics}
    &
    \cite{du2018learning}
    &
    &
    \cite{diligenti2017integrating, stewart2017label, xu2017semantic}
    &
    \cite{ma2018multi,che2015deep,bian2014knowledge}
    &
    \cite{Campos2007bayesian,messaoud2009integrating, heckerman1995learning}
    &
    \cite{choo2013utopian,knox2009interactively, kaplan2017beating}
    \\
    Algorithm
    &
    \cite{heese2019good, fung2003knowledge, mangasarian2008nonlinear}
    &
    \cite{zhu2019}
    &
    &
    &
    \cite{diligenti2017semantic, hu2016deep, hu2016harnessing}
    &
    &
    \cite{borboudakis2012incorporating,feelders2006learning,richardson2003learning}
    &
    \cite{christiano2017deep,brown2012dis,schramowski2020right}
    \\
    &
    \cite{vonkurnatowski2020compensating}
    &
    &
    &
    &
    \cite{chang2007guiding}
    &
    &
    &
    \cite{fails2003interactive, rieger2019interpretations}
    \\
    \midrule
    Final
    &
    \cite{karpatne2017physics, king2018deep}
    &
    &
    \cite{hautier2010finding, du2018learning, pfrommer2018optimisation}
    &
    &
    &
    \cite{glavavs2018explicit, mrkvsic2016counter, fang2017object}
    &
    \cite{yet2014not}
    &
    \\
    Hypothesis
    &
    &
    &
    &
    &
    &
    \cite{vonrueden2020streetmap}
    &
    &
    \\
    &
    &
    &
    &
    &
    &
    &
    &
    \\
    \bottomrule
    \end{tabularx}
\label{tab:rep2int}
\end{table*}

%% file: 5c_simulations.tex
\subsection{Simulation Results}
\label{sec:rep_sim_res}

Simulation results are also a prominent knowledge representation in informed machine learning. They mainly come from scientific knowledge and are used to extend the training data.

\begin{figure}[!h]
    \centering
    \includegraphics{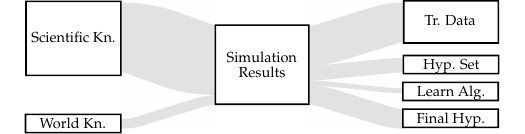}
\end{figure} 

\subsubsection{(Paths from) Knowledge Source}

Computer simulations have a long tradition in many areas of the sciences. While they are also gaining popularity in other domains, most works on integrating simulation results into machine learning deal with natural sciences and engineering.

\textbf{Scientific Knowledge.}
Simulation results informing machine learning can be found in fluid- and thermodynamics~\cite{karpatne2017physics}, material sciences~\cite{hautier2010finding, pfrommer2018optimisation, kim2007hybrid}, life sciences~\cite{deist2019simulation}, mechanics and robotics~\cite{lerer2016learning, rai2019using, du2018learning}, or autonomous driving~\cite{lee2019spigan}.
To make it more concrete, we give three examples:
In material sciences, a density functional theory ab-initio simulation can be used to model the energy and stability of potential new material compounds and their crystal structure~\cite{hautier2010finding}.
Even complex material forming processes can be simulated, for example a composite textile draping process can be simulated based on a finite-element model~\cite{pfrommer2018optimisation}.
As an example for autonomous driving, urban traffic scenes under specific weather and illumination conditions, which might be useful for the training of visual perception components, can be simulated with dedicated physics engines~\cite{lee2019spigan}.

\subsubsection{(Paths to) Knowledge Integration}

We find that the integration of simulation results into machine learning is most often happens via the augmentation of training data. Other approaches that occur frequently are the integration into the hypothesis set or the final hypothesis.

\begin{insertbox}[label={in:simres},floatplacement=t]{Simulation Results as Synthetic Tr. Data}
The results from a simulation can be used as synthetic training data and can thus augment the original, real training data, as illustrated in Figure~\ref{fig:in_simres}. Some papers that follow this approach are~\cite{karpatne2017physics, deist2019simulation, pfrommer2018optimisation, lerer2016learning, shrivastava2017learning, lee2019spigan, rai2019using}.
\\\\
\begin{minipage}[b]{\linewidth}
\centering
\includegraphics{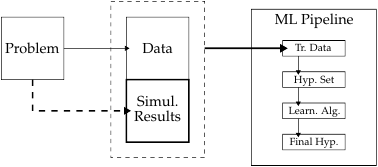}
\captionof{figure}{Information flow for synthetic training data from simulations.}
\label{fig:in_simres}
\end{minipage}

\end{insertbox}

\textbf{Training Data.}
The integration of simulation results into training data~\cite{karpatne2017physics, deist2019simulation, pfrommer2018optimisation, lerer2016learning, shrivastava2017learning, lee2019spigan, rai2019using} depends on how the simulated, i.e. synthetic, data is combined with the real-world measurements: 

Firstly, additional input features are simulated and, together with real data, form input features. For example, original features can be transformed by multiple approximate simulations and the similarity of the simulation results can be used to build a kernel~\cite{deist2019simulation}.

Secondly, additional target variables are simulated and added to the real data as another feature. This way the model does not necessarily learn to predict targets, e.g. an underlying physical process, but rather the systematic discrepancy between simulated and the true target data~\cite{karpatne2017physics}.

Thirdly, additional target variables are simulated and used as synthetic labels, which is of particular use when the original experiments are very expensive~\cite{pfrommer2018optimisation}.
This approach can also be realized with physics engines, for example, pre-trained neural networks
can be tailored towards an application through additional training on simulated data \cite{lerer2016learning}. Synthetic training data generated from simulations can also be used to pre-train components of Bayesian optimization frameworks~\cite{rai2019using}.

In informed machine learning, training data thus stems from a hybrid information source and contains both simulated and real data points (see Insert~\ref{in:simres}).
The gap between the synthetic and the real domain can be narrowed via adversarial networks such as SimGAN. These improve the realism of, say, synthetic images and can generate large annotated data sets by simulation~\cite{shrivastava2017learning}.
The SPIGAN framework goes one step further and uses additional, privileged information from internal data structures of the simulation in order to foster unsupervised domain adaption of deep networks~\cite{lee2019spigan}.

\textbf{Hypothesis Set.}
Another approach we observed integrates simulation results into the hypothesis set~\cite{wang1997knowledge, leary2003knowledge, kim2007hybrid}, which is of particular interest when dealing with low-fidelity simulations. These are simplified simulations that approximate the overall behaviour of a system but ignore intricate details for the sake of computing speed.

When building a machine learning model that reflects the actual, detailed behaviour of a system, low-fidelity simulation results or a response surface (a data-driven model of the simulation results) can be build into the architecture of a knowledge-based neural network (KBANN~\cite{towell1994knowledge}, see Insert~\ref{in:kbann}), e.g.~by replacing one or more neurons.
This way, parts of the network can be used to learn a mapping from low-fidelity simulation results to a few real-world observations or high-fidelity simulations~\cite{leary2003knowledge, kim2007hybrid}.

\textbf{Learning Algorithm.}
Furthermore, a simulation can directly be integrated into iterations of a a learning algorithm. For example, a realistic positioning of objects in a 3D scene can be improved by incorporating feedback from a solid-body simulation into learning~\cite{du2018learning}. By means of reinforcement learning, this is even feasible if there are no gradients available from the simulation.

\textbf{Final Hypothesis.} A last but important approach that we found in our survey integrates simulation results into the final hypothesis set of a machine learning model. Specifically, simulations can validate results of a trained model~\cite{hautier2010finding, pfrommer2018optimisation, du2018learning}.

%% file: 5d_local_invariances.tex
\subsection{Spatial Invariances}
\label{sec:rep_inv}

Next, we describe informed machine learning approaches involving the representation type of spatial invariances.
Their main path comes from world knowledge and goes to the hypothesis set.

\begin{figure}[!h]
    \centering
    \includegraphics{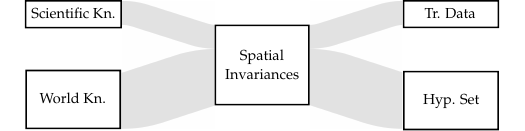}
\end{figure} 

\subsubsection{(Paths from) Knowledge Source}

We mainly found references using spatial invariances in the context of world knowledge or scientific knowledge.

\textbf{World Knowledge.}
Knowledge about invariances may fall into the category of world knowledge, for example when modeling facts about local or global pixel correlations in images \cite{scholkopf1998prior}. Indeed, invariants are often used in image recognition where many characteristics are invariant under metric-preserving transformations. For example, in object recognition, an object should be classified correctly independent of its rotation in an image. 

\textbf{Scientific Knowledge.}
In physics, Noether's theorem states that certain symmetries (invariants) lead to conserved quantities (first integrals) and thus integrate Hamiltonian systems or equations of motion \cite{wu2018physics}, \cite{ling2016reynolds}. For example, in equations modeling planetary motion, the angular momentum serves as such an invariant.

\subsubsection{(Paths to) Knowledge Integration}

In most references we found spatial invariances informing the hypothesis set.

\textbf{Hypothesis Set.}
Invariances from physical laws can be integrated into the architecture of a neural network. For example, invariant tensor bases can be used to embed Galilean invariance for the prediction of fluid anisotropy tensors~\cite{ling2016reynolds}, or the physical Minkowski metric that reflects mass invariance can be integrated via a Lorentz layer into a neural network~\cite{butter2018deep}.

A recent trend is to integrate knowledge as spatial invariances into the architecture or layout of convolutional neural networks, which leads to so called geometric deep learning in~\cite{bronstein2017geometric}. 
A natural generalization of CNNs are group equivariant CNNs (G-CNNs)~\cite{cohen2016group, dieleman2016exploiting, li2018deep}. G-convolutions provide a higher degree of weight sharing and expressiveness. Simply put, the idea is to define filters based on a more general group-theoretic convolution. 
Another approach towards rotation invariance in image recognition considers harmonic network architecture where a certain response entanglement (arising from features that rotate at different frequencies) is resolved \cite{worrall2017harmonic}. The goal is to design CNNs that exhibits equivariance to patch-wise translation and rotation by replacing conventional CNN filters with circular harmonics.

In support vector machines, invariances under group transformations and prior knowledge about locality can be incorporated by the construction of appropriate kernel functions \cite{scholkopf1998prior}. In this context, local invariance is defined in terms of a regularizer that penalizes the norm of the derivative of the decision function \cite{lauer2008incorporating}. 

\textbf{Training Data.}
An early example of integrating knowledge as invariances into machine learning is the creation of virtual examples \cite{niyogi1998incorporating} and it has been shown that data augmentation through virtual examples is mathematically equivalent to incorporating prior knowledge via a regularizer.
A similar approach is the creation of meta-features~\cite{bergman2019symmetry}. For instance, in turbulence modelling using the Reynolds stress tensor, a feature can be createad that is rotational, reflectional and Galilean invariant \cite{wu2018physics}. This is achieved by selecting features fulfilling rotational and Gallilean symmetries and augmenting the training data to ensure reflectional invariance.

%% file: 5e_logic_rules.tex
\subsection{Logic Rules}
\label{sec:rep_log_rul}

Logic Rules play an important role for the integration of prior knowledge into machine learning.
In our literature survey, we mainly found the the source of world knowledge and the two integration paths into the hypothesis set and the learning algorithm.

\begin{figure}[!h]
    \centering
    \includegraphics{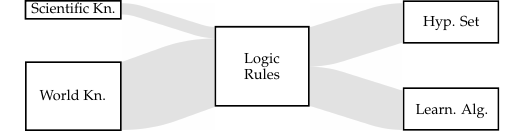}
\end{figure} 

\subsubsection{(Path from) Knowledge Source}

Logic rules can formalize knowledge from various sources, but the most frequent is world knowledge. Here we give some illustrative examples.

\textbf{World Knowledge.}
Logic rules often describe knowledge about real-world objects~\cite{stewart2017label, diligenti2017integrating, xu2017semantic, sachan2018learning, schiegg2012markov} such as seen in images.
This can focus on object properties, such as for animals $x$ that $(\text{FLY}(x) \land \text{LAYEGGS}(x) \Rightarrow \text{BIRD}(x))$~\cite{diligenti2017integrating}.
It can also focus on relations between objects such as the co-occurrence of characters in game scenes, e.g.~$(\text{PEACH} \Rightarrow \text{MARIO}$)~\cite{stewart2017label}. 

Another knowledge domain that can be well represented by logic rules is linguistics~\cite{hu2016harnessing, hu2016deep, chang2007guiding, chang2012structured, kimmig2012short, richardson2006markov, sridhar2015joint}.
Linguistic rules can consider the sentiment of a sentence (e.g., if a sentence consists of two sub-clauses connected with a 'but', then the sentiment of the clause after the 'but' dominates~\cite{hu2016harnessing}); or the order of tags in a given word sequence (e.g., if a given text element is a citation, then it can only start with an author or editor field~\cite{chang2007guiding}).

Rules can also describe dependencies in social networks. For example,
on a scientific research platform, it can be observed that authors citing each other tend to work in the same field (Cite$(x, y)$ $\land$ hasFieldA$(x)$ $\Rightarrow$ hasFieldA$(y)$) \cite{diligenti2017semantic}.

\subsubsection{(Path to) Knowledge Integration}

We observe that logic rules are integrated into learning mainly in the hypothesis set or, alternatively, in the learning algorithm.

\textbf{Hypothesis Set.}
Integration into the hypothesis set comprises both deterministic and probabilistic approaches. The former include neural-symbolic systems, which use rules as the basis for the model structure~\cite{towell1994knowledge, garcez1999connectionist, francca2014fast}.
In Knowledge-Based Artificial Neural Networks (KBANNs), the architecture is constructed from symbolic rules by mapping the components of propositional rules to network components~\cite{towell1994knowledge} as further explained in Insert~\ref{in:kbann}.
Extensions are available
that also output a revised rule set~\cite{garcez1999connectionist}
or also consider first-order logic~\cite{francca2014fast}.
A recent survey about neural-symbolic computing~\cite{garcez2019neural} summarizes further methods.

\begin{insertbox}[label={in:kbann},floatplacement=t]{Knowledge-Based Artificial Neural Networks (KBANNs)}
Rules can be integrated into neural architectures by mapping the rule's components to the neurons and weights with these steps illustrated in Figure~\ref{fig:kbann}~\cite{towell1994knowledge}:
\begin{enumerate}
    \item Get rules. If needed, rewrite them to have a hierarchical structure.
    \item Map rules to a network architecture. Construct (positively/negatively) weighted links for (existing/negated) dependencies.
    \item Add nodes. These are not given through the initial rule set and represent hidden units.
    \item Perturb the complete set of weights.
\end{enumerate}
After the KBANN's architecture is built, the network is refined with learning algorithms.
\\\\
\begin{minipage}[b]{\linewidth}
\centering
\includegraphics{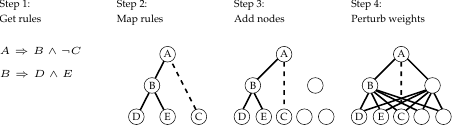}
\captionof{figure}{Steps of Rules-to-Network Translation~\cite{towell1994knowledge}. Simple example for integrating rules into a KBANN.}
\label{fig:kbann}
\end{minipage}
\end{insertbox}

Integrating logic rules into the hypothesis set in a probabilistic manner is yet another approach \cite{richardson2006markov, kimmig2012short, schiegg2012markov, sachan2018learning}.
These belong to the research direction of statistical relational learning~\cite{Raedt2016Statistical}.
Corresponding frameworks provide a logic templating language to define a probability distribution over a set of random variables.
Two prominent frameworks are markov logic networks~\cite{richardson2006markov, schiegg2012markov} and probabilistic soft logic~\cite{kimmig2012short, sachan2018learning}, which translate a set of first-order logic rules to a markov random field. Each rule specifies dependencies between random variables and serves as a template for so called potential functions, which assign probability mass to joint variable configurations.

\textbf{Learning Algorithm.}
The integration of logic rules into the learning algorithm is often accomplished via additional, semantic loss terms~\cite{diligenti2017integrating, diligenti2017semantic, hu2016harnessing, hu2016deep, stewart2017label, xu2017semantic, chang2007guiding}.
These augment the objective function similar to the knowledge-based loss terms explained above.
However, for logic rules, the additional loss terms evaluate a functional that transforms rules into continuous and differentiable constraints, for example via the t-norm~\cite{diligenti2017integrating}.
Semantic loss functions can also be derived from first principles using a set of axioms~\cite{xu2017semantic}.
As a specific approach for student-teacher architectures, the rules can be first integrated in a teacher network and can then be used by a student network that is trained by minimizing a semantic loss term that measures the imitation of the teacher network~\cite{hu2016harnessing, hu2016deep}.

%% file: 5f_knowledge_graphs.tex
\subsection{Knowledge Graphs}
\label{graphs}
\label{sec:rep_kno_gra}

The taxonomy paths we observed in our literature survey that are related to knowledge representation are illustrated in the following graphic.

\begin{figure}[!h]
    \centering
    \includegraphics{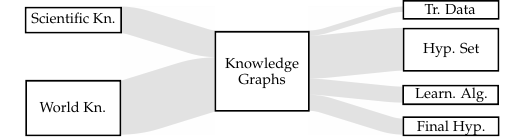}
\end{figure} 

\subsubsection{(Paths from) Knowledge Source}

Since graphs are very versatile modeling tools, they can represent various kinds of structured knowledge. Typically, they are constructed from databases, however, the most frequent source we found in informed machine learning papers is world knowledge.

\textbf{World Knowledge.}
Since humans perceive the world as composed of entities, graphs are often used to represent relations between visual entities. For example, the Visual Genome knowledge graph is build from human annotations of object attributes and relations between objects in natural images \cite{marino2017more, jiang2018hybrid}. Similarly, the MIT ConceptNet \cite{speer2013conceptnet} encompasses concepts of everyday life and their relations automatically built from text data.
In natural language processing, knowledge graphs often represent knowledge about relations among concepts, which can be referred to by words. For example, WordNet~\cite{miller1995wordnet} represents semantic and lexical relations of words such as synonymy. 
Such knowledge graphs are often used for information extraction in natural language processing, but information extraction can also be used to build new knowledge graphs~\cite{mitchell2018never}.

\textbf{Scientific Knowledge.}
In physics, graphs can immediately describe physical systems such as spring-coupled masses \cite{battaglia2016interaction}.
In medicine, networks of gene-protein interactions describe biological pathway information \cite{ma2018multi} and the hierarchical nature of medical diagnoses is captured by classification systems such as the International Classification of Diseases (ICD) \cite{che2015deep, choi2017gram}.

\subsubsection{(Paths to) Knowledge Integration}

In our survey, we observed the integration of knowledge graphs in all four components of the machine learning pipeline but most prominently in the hypothesis set.

\begin{insertbox}[label={in:graph},floatplacement=t]{Integrating Knowledge Graphs in CNNs for Image Classification}
Image classification through convolutional neural networks can be improved by using knowledge graphs that reflect relations between detected objects, as illustrated in Figure~\ref{fig:graphinsert}.
Technically, such relations form adjacency matrices in gated graph neural networks~\cite{marino2017more}. During the detection, the network graph is propagated, starting with detected nodes and then expanding to neighbors~\cite{battaglia2018relational}.
\\\\
\begin{minipage}[b]{\linewidth}
\centering
\includegraphics{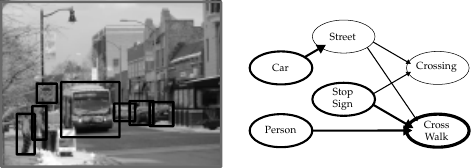}
\captionof{figure}{Illustrative application example of using neural networks and knowledge graphs for image classification, similar as in~\cite{marino2017more}. The image (from the COCO dataset) shows a pedestrian cross walk.}
\label{fig:graphinsert}
\end{minipage}
\end{insertbox}

\textbf{Hypothesis Set.}
The fact that the world consists of interrelated objects can be integrated by altering the hypothesis set. Graph neural networks operate on graphs and thus feature an object- and relation-centric bias in their architecture \cite{battaglia2018relational}. A recent survey \cite{battaglia2018relational} gives an overview over this field and explicitly names this knowledge integration relational inductive bias. This bias is of benefit, e.g. for learning physical dynamics~\cite{battaglia2016interaction, chang2016compositional} or object detection~\cite{jiang2018hybrid}.

In addition, graph neural networks allow for the explicit integration of a given knowledge graph as a second source of information. This allows for multi-label classification in natural images where inference about a particular object is facilitated by using relations to other objects in an image~\cite{marino2017more} (see Insert~\ref{in:graph}). More generally, a graph reasoning layer can be inserted into any neural network~\cite{liang2018symbolic}. The main idea is to enhance representations in a given layer by propagating through a given knowledge graph.

Another approach is to use attention mechanisms on a knowledge graph in order to enhance features. In natural language analysis, this facilitates the understanding as well as the generation of conversational text~\cite{zhou2018commonsense}. Similarly, graph-based attention mechanism are used to counteract too few data points by using more general categories \cite{choi2017gram}.
Also, attention on related knowledge graph embedding can support the training of word embeddings like ERNIE~\cite{zhang2019ernie}, which are fed into language models like BERT~\cite{devlin2018bert, peters2019knowledge}.
 
\textbf{Training Data.}
Another prominent approach is distant supervision where information in a graph is used to automatically annotate texts to train natural language processing systems. This was originally done na\"ively by considering each sentence that matches related entities in a graph as a training sample \cite{mintz2009distant}; however, recently attention-based networks have been used to reduce the influence of noisy training samples \cite{ye2019distant}.

\textbf{Learning Algorithm.}
Various works discuss the integration of graph knowledge into the learning algorithm. For instance, a regularization term based on the graph Laplacian matrix can enforce strongly connected variables to behave similarly in the model, while unconnected variables are free to contribute differently. This is commonly used in bioinformatics to integrate genetic pathway information \cite{ma2018multi,che2015deep}. 
Some natural language models, too, include information from a knowledge graph into the learning algorithm, e.g.~when computing word embeddings. Known relations among words can be utilized as augmented contexts \cite{bian2014knowledge} in word2vec training \cite{mikolov2013efficient}.

\textbf{Final Hypothesis.}
Finally, graph can also be used to improve or validate final hypotheses or trained models. For instance, a recent development is to post-process word embeddings based on information from knowledge graphs \cite{mrkvsic2016counter, glavavs2018explicit}.
Furthermore, semantic segmentation in autonomous driving can be validated using knowledge graphs of street maps~\cite{vonrueden2020streetmap},
or in object detection, predicted probabilities of a learning system can be refined using semantic consistency measures \cite{fang2017object} derived form knowledge graphs. In both cases, the knowledge graphs are used to indicate whether the prediction is consistent with available knowledge.

%% file: 5g_probabilisitic_relations.tex
\subsection{Probabilistic Relations}
\label{sec:rep_pro_rel}

The most frequent paths probabilistic relations found in our literature survey comes from expert knowledge and goes to the hypothesis set or the learning algorithm. 

\begin{figure}[!h]
    \centering
    \includegraphics{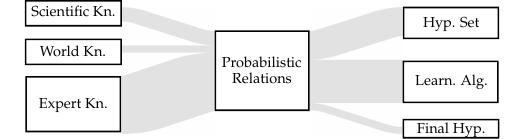}
\end{figure}

\subsubsection{(Paths from) Knowledge Source}

Knowledge in form of probabilistic relations originates most prominently from domain experts, but can also come from other sources such as natural sciences. 

\textbf{Expert Knowledge.}
A human expert has intuitive knowledge over a domain, for example, which entities are related to each other and which are independent. Such relational knowledge, however, is often not quantified and validated and differs from, say, knowledge in natural sciences. Rather, it involves degrees of belief or uncertainty.

Human expertise exists in all domains.
In the car insurance, driver features like age relate to risk aversion \cite{Campos2007bayesian}.
Another examples is computer expertise for troubleshooting, i.e relating a device status to observations \cite{richardson2006markov}. 

\textbf{Scientific Knowledge.}
Correlation structures can also be obtained from natural sciences knowledge. For example, correlations between genes can be obtained from gene interaction networks \cite{massa2010gene} or from a gene ontology \cite{messaoud2009integrating}.

\subsubsection{(Paths to) Knowledge Integration}

We generally observe the integration of probabilistic relations into the hypothesis set as well as into the learning algorithm and the final hypothesis.

\textbf{Hypothesis Set.}
Expert knowledge is the basis for probabilistic graphical models. For example, Bayesian network structures are typically designed by human experts and thus fall into the category of informing the hypothesis set.
Here, we focus on contributions where knowledge and Bayesian inference are combined in more intricate ways, for instance, by learning network structures from knowledge and from data. A recent overview \cite{angelopoulos2008bayesian} categorizes the type of prior knowledge about network structures into the presence or absence of edges, edge probabilities, and knowledge about node orders.

Probabilistic knowledge can be used directly in the hypothesis set. For example,
extra nodes can be added to a Bayesian network thus altering the hypothesis set \cite{constantinou2016integrating}, or the structure of a probabilistic model can be chosen in accordance to given spatio-temporal structures~\cite{piatkowski2013spatio}.
In other hybrid approaches, the parameters of the conditional distribution of the Bayesian network are either learned from data or obtained from knowledge \cite{yet2014combining, heckerman1995learning}. 

\textbf{Learning Algorithm.}
Human knowledge can also be used to define an informative prior \cite{heckerman1995learning, fischer20no}, which affects the learning algorithm as is has a regularizing effect.
Structural constraints can alter score functions or the selection policies of conditional independence test, informing the search for the network structure \cite{Campos2007bayesian}.
More qualitative knowledge, e.g. observing one variable increases the probability of another, was integrated using isotonic regression, i.e. parameter estimation with order constraints \cite{feelders2006learning}.
Causal network inference can make use of ontologies to select the tested interventions \cite{messaoud2009integrating}.
Furthermore, prior causal knowledge can be used to constrain the direction of links in a Bayesian network \cite{borboudakis2012incorporating}.

\textbf{Final Hypothesis.}
Finally, predictions obtained from a Bayesian network can be judged by probabilistic relational knowledge in order to refine the model \cite{yet2014not}.

%% file: 5h_human_feedback.tex
\subsection{Human Feedback}
\label{sec:rep_hum_fee}

Finally, we look at informed machine learning approaches belonging to the representation type of human feedback.
The most common path begins with expert knowledge and ends at the learning algorithm.

\begin{figure}[!h]
    \centering
    \includegraphics{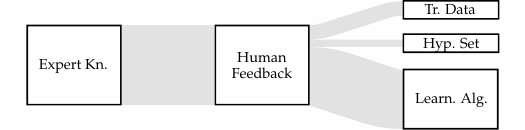}
\end{figure} 

\subsubsection{(Paths from) Knowledge Source}
Compared to other categories in our taxonomy, knowledge representation via human feedback is less formalized and mainly stems from expert knowledge. 

\textbf{Expert Knowledge.}
Examples of knowledge that fall into this category include knowledge about topics in text documents \cite{choo2013utopian}, agent behaviors \cite{knox2009interactively, kaplan2017beating, christiano2017deep, hester2018deep}, and data patterns and hierarchies \cite{ choo2013utopian, brown2012dis, schramowski2020right}.
Knowledge is often provided in form of relevance or preference feedback and humans in the loop can integrate their intuitive knowledge into the system without providing an explanation for their decision. 
For example, in object recognition, users can provide their corrective feedback about object boundaries via brush strokes \cite{fails2003interactive}.
As another example, in Game AI, an expert user can give spoken instructions for an agent in an Atari game~\cite{kaplan2017beating}.

\subsubsection{(Paths to) Knowledge Integration}

Human feedback for machine learning is usually assumed to be limited to feature engineering and data annotation. However, it can also be integrated into the learning algorithm itself.
This often occurs in areas of reinforcement learning, or interactive learning combined with visual analytics.

\textbf{Learning Algorithm.}
In reinforcement learning, an agent observes an unknown environment and learns to act based on reward signals. The TAMER framework \cite{knox2009interactively} provides the agent with human feedback rather than (predefined) rewards. This way, the agent learns from observations and human knowledge alike.
While these approaches can quickly learn optimal policies, it is cumbersome to obtain the human feedback for every action. Human preference w.r.t.~whole action sequences, i.e. agent behaviors, can circumvent this \cite{christiano2017deep}. This enables the learning of reward functions.
Expert knowledge can also be incorporated through natural language interfaces\cite{kaplan2017beating}. Here, a human provides instructions and agents receive rewards upon completing these instructions.

Active learning offers a way to include the ``human in the loop'' to efficiently learn with minimal human intervention. This is based on iterative strategies where a learning algorithm queries an annotator for labels \cite{settles2009active}. We do not consider this standard active learning as an informed learning method because the human knowledge is essentially used for label generation only.
However, recent efforts integrate further knowledge into the active learning process.

Visual analytics combines analysis techniques and interactive visual interfaces to enable exploration of --and inference from-- data~\cite{keim2008visual}. Machine learning is increasingly combined with visual analytics.
For example, visual analytics systems allow users to
drag similar data points closer in order to learn distance functions~\cite{brown2012dis},
provide corrective feedback in object recognition \cite{fails2003interactive},
or even to alter correctly identified instances where the interpretation is not in line with human explanations~\cite{schramowski2020right, rieger2019interpretations}.

Lastly, various tools exist for text analysis, in particular for topic modeling \cite{choo2013utopian} where users can create, merge and refine topics or change keyword weights. They thus impart knowledge by generating new reference matrices (term-by-topic and topic-by-document matrices) that are integrated in a regularization term that penalizes the difference between the new and the old reference matrices. This is similar to the semantic loss term described above. 

\textbf{Training Data and Hypothesis Set.}
Another approach towards incorporating expert knowledge in reinforcement learning considers human demonstration of problem solving. Expert demonstrations can be used to pre-train a deep Q-network, which accelerates learning \cite{hester2018deep}. Here, prior knowledge is integrated into the hypothesis set and the training data since the demonstrations inform the training of the Q-network and, at the same time, allow for interactive learning via simulations.

%% file: 6_related_work.tex
\section{Historical Background}
\label{sec:background}

The idea of integrating knowledge into learning has a long history. 
Historically, AI research roughly considered the two antipodal paradigms of symbolism and connectionism. The former dominated up until the 1980s and refers to reasoning based on symbolic knowledge; the latter became more popular in the 1990s and considers data-driven decision making using neural networks.
Especially Minsky~\cite{minsky1991logical} pointed out limitations of symbolic AI and promoted a stronger focus on data-driven methods to allow for causal and fuzzy reasoning. Already in the 1990s were knowledge data bases used together with training data to obtain knowledge-based artificial neural networks \cite{towell1994knowledge}.
In the 2000s, when support vector machines (SVMs) were the de-facto paradigm in classification, there was interest in incorporating knowledge into this formalism \cite{lauer2008incorporating}. Moreover, in the geosciences, and most prominently in weather forecasting, knowledge integration dates back to the 1950s. Especially the discipline of data assimilation deals with techniques that combine statistical and mechanistic models to improve prediction accuracy~\cite{kalnay2003atmospheric,reich2015probabilistic}.

%% file: 7_discussion.tex
\section{Discussion of Challenges and Directions}
\label{sec:discussion}

\input{7x_table}

Our findings about the main approaches of informed machine learning are summarized in Table~\ref{tab:main_approaches}. It gives for each approach the taxonomy path, its main motivation, the central approach idea, remarks to potential challenges, and our viewpoint on current or future directions.
For further details on the methods themselves and the corresponding papers, we refer to Section~\ref{sec:approaches}.
In the following, we discuss the challenges and directions for these main approaches, sorted by the integrated knowledge representations.

Prior knowledge in the form of algebraic equations can be integrated as constraints via knowledge-based loss terms (e.g., \cite{karpatne2017physics, stewart2017label, heese2019good}).
Here, we see a potential challenge in
finding the right weights for supervision from knowledge vs. data labels.
Currently, this is solved by setting the hyperparameters for the individual loss terms~\cite{karpatne2017physics}.
However, we think that strategies from more recently developed learning algorithms, such as self-supervised~\cite{janner2017self} or few-shot learning~\cite{wang2020generalizing},
could also advance the supervision from prior knowledge.
Moreover, we suggest further research on theoretical concepts based on the existing generalization bounds from statistical learning theory~\cite{cucker2007learning, steinwart2008support} and the connection between regularization and effective hypothesis space~\cite{cucker2002best}.

Differential equations can be integrated similarly, but with a specific focus on physics-informed neural networks that constrain the model derivatives by the underlying differential equation (e.g., \cite{lagaris1998artificial, raissi2017physics1, zhu2019}).
A potential challenge is the robustness of the solution, which is the subject of current research. One approach is to investigate the the model quality by a suitable quanitification of its uncertainty~\cite{yang2018physics, zhu2019}.
We think, a more in-depth comparison with
existing numerical solvers~\cite{lapidus2011numerical} would also be helpful.
Another challenge of physical systems is the generation and integration of sensor data in real-time.
This is currently tackled by online learning methods~\cite{lutter2019deep}.
Furthermore, we think that techniques from data assimilation~\cite{reich2015probabilistic} could also be helpful to combine modelling from knowledge and data.

Simulation results can be used for synthetic data generation or augmentation (e.g., \cite{deist2019simulation, pfrommer2018optimisation, lee2019spigan}),
but this can bring up the challenge of a mismatch between real and simulated data.
A promising direction to close the gap is domain adaptation, especially adversarial training~\cite{shrivastava2017learning, wulfmeier2017addressing}, or domain randomization~\cite{peng2018sim}.
Moreover, for future work we see further potential in the development of new hybrid systems that combine machine learning and simulation in more sophisticated ways~\cite{vonrueden2020combining}.

The utilization of spatial invariances through model architectures with invariant characteristics, such as group equivariant or convolutional networks, diminish the model search space (e.g., \cite{cohen2016group, dieleman2016exploiting, worrall2017harmonic}).
Here, a potential challenge is the proper invariance specification and implementation~\cite{worrall2017harmonic} or expensive evaluations on more complex geometries~\cite{bronstein2017geometric}.
Therefore, we think that the efficient adaptation of invariant-based models to further scenarios can further improve geometric-based representation learning~\cite{bronstein2017geometric}. 

Logic rules can be encoded in the architecture of knowledge-based neural networks (KBANNs), (e.g., \cite{towell1994knowledge, garcez1999connectionist, francca2014fast}).
Since this idea was already developed when neural networks had only a few layers, a question is, if it is still feasible for deep neural networks.
In order to improve the practicality, we suggest to develop automated interfaces for knowledge integration.
A future direction could be the development of new neuro-symbolic systems. Although the combination of connectionist and symbolic systems into hybrid systems is a longtime idea~\cite{mcgarry1999hybrid, sun2006connectionist}, it is currently getting more attention~\cite{garcez2020neurosymbolic, dong2018imposing}.
Another challenge, especially in statistical relational learning (SRL), such as Markov logic networks or probabilistic soft logic (e.g., \cite{kimmig2012short, sachan2018learning, bach2015Hinge}).
is the aquisition of rules when they are not yet given.
An ongoing research topic to this end is the learning of rules from data, which is called structure learning~\cite{Embar2018ScalableSL}.

Knowledge graphs can be integrated into learning systems either explicitly via graph propagation and attention mechanisms, or implicitly via graph neural networks with relational inductive bias (e.g., \cite{battaglia2016interaction, jiang2018hybrid, marino2017more}).
A challenge is the comparability between different methods, because authors often use template like ConceptNet~\cite{zhou2018commonsense} or VisualGenome~\cite{marino2017more, jiang2018hybrid} and customize the graphs in to improve running time and performance.
Since the choice of graph can have high influence~\cite{liang2018symbolic}, we suggest a pool of standardized graphs in order to improve comparability, or even to establish benchmarks.
Another interesting direction is to combine graph using and graph learning.
A requirement here is the need for good entity linking models in approaches such as KnowBERT~\cite{peters2019knowledge} and ERNIE~\cite{zhang2019ernie} and the continuous embedding of new facts in the graph.

Probabilistic Relations can be integrated as prior knowledge in terms of a-priori probability distributions that are refined with additional observations (e.g.,~\cite{yet2014combining, heckerman1995learning, constantinou2016integrating}).
The main challenges are the large computational effort and the formalization of knowledge in terms of inductive priors.
Directions responding to this
are variational methods with origins in optimization theory and functional analysis~\cite{blei2017variational} and variational
neural networks~\cite{kingma2019introduction}. Besides scaling issues, an explicit treatment of causality is becoming more important in machine learning and closely related to graphical probabilistic models~\cite{pearl2009causality}.

Human feedback can be integrated into the learning algorithm by
human-in-the-loop (HITL) reinforcement learning (e.g.,~\cite{knox2009interactively, christiano2017deep}),
or by
explanation alignment through interactive learning combined with visual analytics (e.g.,~\cite{rieger2019interpretations, schramowski2020right}).
However, the exploration of human feedback can be very expensive due to its latency in real systems.
Exploratory actions could hamper user experience~\cite{kreutzer2020learning, dulac2019challenges}, so that online reinforcement learning is generally avoided.
A promising approach is learning a reward estimator~\cite{kreutzer2018reliability, gao2018april} from collected logs, which then provides unlimited feedback for unseen instances that do not have any human judgments.
Another challenge is that human feedback is often intuitive and not formalized and thus difficult to incorporate into machine learning systems. Also human-gorunded evaluation is very costly, especially compared to functionally-grounded evaluation~\cite{doshi2017towards}.
Therefore we suggest to further study representation transformations to formalize intuitive knowledge, e.g. from human feedback to logical rules. 
Furthermore,
we found that improved interpretability still only is a minor goal for knowledge integration (see Figure~\ref{fig:integration_goals}). This, too, suggests opportunities for future work.

Even if these directions are motivated by specific approaches, we think that they are generally relevant and can advance the whole field of informed machine learning.

%% file: 7x_table.tex
\newcolumntype{Y}{>{\raggedright\arraybackslash}X}


\begin{table*}
\caption{
\textbf{Main Approaches of Informed Machine Learning.}
The approaches are sorted by taxonomy path and knowledge representation.
Methodical details can be found in Section~\ref{sec:approaches}.
Challenges and directions are discussed in Section~\ref{sec:discussion}.
}
\renewcommand{\arraystretch}{1.3}
\scriptsize
\centering
    \begin{tabularx}{\textwidth}{@{}lllllll@{}}
    \toprule
    \multicolumn{3}{l}{\textbf{Taxonomy Path}}
    &
    \multicolumn{1}{l}{\textbf{Main Motivation}}
    &
    \multicolumn{1}{l}{\textbf{Central Approach Idea}}
    &
    \multicolumn{1}{l}{\textbf{Potential Challenge}}
    &
    \multicolumn{1}{l}{\textbf{Current / Future Directions}}
    \\
    \cmidrule{1-3}
    Source&
    Represent.&
    Integration&
    &
    &
    &
    \\
    \midrule
    Scientific
    &
    Algebraic
    &
    Learning
    &
    Less data,
    &
    Knowledge-based loss
    &
    Weighting supervision
    &
    Hyperparameter setting, 
    \\
    Knowl.
    &
    Equations
    &
    Algor.
    &
    Knowl. conform.
    &
    terms from constraints
    &
    from data labels vs.
    &
    Novel learning algorithms,
    \\
    &
    \tiny{(See Sec.~\ref{sec:rep_alg_eqn})}
    &
    &
    &
    (see Insert~\ref{in:loss})
    &
    knowledge
    &
    Extension of learning theory
    \\
    \cmidrule{2-7}
    &
    Differential
    &
    Learning
    &
    Knowl. conform.,
    &
    Physics-informed neural
    &
    Solution robustness,
    &
    Uncertainty quantification,
    \\
    &
    Equations
    &
    Algor.
    &
    Less data
    &
    networks with derivatives
    &
    Real-time data generation
    &
    Numerical solver comparison,
    \\
    &
    \tiny{(See Sec.~\ref{sec:rep_dif_eqn})}
    &
    &
    &
    in loss function
    &
    and integration
    &
    Online learning, data assimilation
    \\
    \cmidrule{2-7}
    &
    Simulation
    &
    Training
    &
    Less data
    &
    Synthetic data generation 
    &
    Sim-to-real gap, i.e.
    &
    Adversarial domain adaptation,
    \\
    &
    Results
    &
    Data
    &
    &
    
    or data augmentation
    &
    mismatch between real
    &
    Domain randomization;
    \\
    &
    \tiny{(See Sec.~\ref{sec:rep_sim_res})}
    &
    &
    &
    (see Insert~\ref{in:simres})
    &
    and simulated data
    &
    Hybrid systems
    \\
    \midrule
    World
    &
    Spatial
    &
    Hypoth.
    &
    Performance
    &
    Models with invariant
    &
    Invariance specification,
    &
    Geometric-based
    \\
    Knowl.
    &
    Invariances
    &
    Set
    &
    (Small models)
    &
    characteristics, e.g. group
    &
    expensive geometric
    &
    representation learning,
    \\
    &
    \tiny{(See Sec.~\ref{sec:rep_inv})}
    &
    &
    &
    equivariant DNNs/CNNs
    &
    evaluations
    &
    Adaptaion to complex scenarios
    \\
    \cmidrule{2-7}
    &
    Logic
    &
    Hypoth.
    &
    Performance
    &
    KBANNs (see Insert~\ref{in:kbann});
    &
    Feasibility for deep
    &
    Automated integration interface,
    \\
    &
    Rules
    &
    Set
    &
    
    &
    SRL (e.g., Markov logic
    &
    neural networks;
    &
    Neuro-symbolic systems;
    \\
    &
    \tiny{(See Sec.~\ref{sec:rep_log_rul})}
    &
    &
    &
    networks, prob. soft logic)
    &
    Acquisition of rules
    &
    Structure learning
    \\
    \cmidrule{2-7}
    &
    Knowl.
    &
    Hypoth.
    &
    Performance,
    &
    Gr. propagation (see Insert~\ref{in:graph}),
    &
    Comparability with custom
    &
    Standardized graph data pool,
    \\
    &
    Graphs
    &
    Set
    &
    Less data
    &
    attention; Gr. neural networks
    &
    graphs, Getting the graph,
    &
    Combine graph using and learning,
    \\
    &
    \tiny{(See Sec.~\ref{sec:rep_kno_gra})}
    &
    &
    &
    (relational inductive bias)
    &
    Entity linking
    &
    Neuro-symbolic systems
    \\
    \midrule
    Expert
    &
    Probabilistic
    &
    Hypoth.
    &
    Less data
    &
    Informed structure of
    &
    High computational
    &
    Variational methods combining
    \\
    Knowl.
    &
    Relations
    &
    Set
    &
    
    &
    prob. graphical models,
    &
    effort, Formalization
    &
    prob. models with numerical opt.,
    \\
    
    &
    \tiny{(See Sec.~\ref{sec:rep_pro_rel})}
    &
    &
    &
    informative priors
    &
    of knowledge
    &
    Probabilistic neural networks
    \\
    \cmidrule{2-7}
    &
    Human
    &
    Learning
    &
    Less data,
    &
    HITL Reinforcement learning;
    &
    Feedback latency;
    &
    Reward estimation from logs;
    \\
    &
    Feedback
    &
    Algor.
    &
    Performance,
    &
    Explanation alignment via
    &
    Formalization of intuition,
    &
    Representation transformation,
    \\
    &
    \tiny{(See Sec.~\ref{sec:rep_hum_fee})} 
    &
    &
    Interpretability
    &
    Visual anal./interactive ml
    &
    Evaluation methods
    &
    Utilization for interpretability
    \\
    \bottomrule
    \end{tabularx}
\label{tab:main_approaches}
\end{table*}


%% file: 8_conclusion.tex
\section{Conclusion}
\label{sec:conclusion}

In this paper, we presented a 
unified classification framework for the explicit integration of additional prior knowledge into machine learning, which we described using the umbrella term of \textit{informed machine learning}.
Our main contribution is the development of a taxonomy that allows a structured categorization of approaches and the uncovering of main paths. Moreover, we presented a conceptual clarification of informed machine learning, as well as a systematic and comprehensive research survey.
This helps current and future users of informed machine learning to identify the right methods to use their prior knowledge, for example, to deal with insufficient training data or to make their models more robust.

%% file: ms.bbl
\begin{thebibliography}{100}
\providecommand{\url}[1]{#1}
\csname url@samestyle\endcsname
\providecommand{\newblock}{\relax}
\providecommand{\bibinfo}[2]{#2}
\providecommand{\BIBentrySTDinterwordspacing}{\spaceskip=0pt\relax}
\providecommand{\BIBentryALTinterwordstretchfactor}{4}
\providecommand{\BIBentryALTinterwordspacing}{\spaceskip=\fontdimen2\font plus
\BIBentryALTinterwordstretchfactor\fontdimen3\font minus
  \fontdimen4\font\relax}
\providecommand{\BIBforeignlanguage}[2]{{%
\expandafter\ifx\csname l@#1\endcsname\relax
\typeout{** WARNING: IEEEtran.bst: No hyphenation pattern has been}%
\typeout{** loaded for the language `#1'. Using the pattern for}%
\typeout{** the default language instead.}%
\else
\language=\csname l@#1\endcsname
\fi
#2}}
\providecommand{\BIBdecl}{\relax}
\BIBdecl

\bibitem{krizhevsky2012imagenet}
A.~Krizhevsky, I.~Sutskever, and G.~E. Hinton, ``Imagenet classification with
  deep convolutional neural networks,'' in \emph{Neural Information Processing
  Systems (NIPS)}, 2012.

\bibitem{hinton2012deep}
G.~Hinton, L.~Deng, D.~Yu, G.~Dahl, A.-R. Mohamed, N.~Jaitly, A.~Senior,
  V.~Vanhoucke, P.~Nguyen, B.~Kingsbury \emph{et~al.}, ``Deep neural networks
  for acoustic modeling in speech recognition,'' \emph{Signal Processing
  Magazine}, vol.~29, 2012.

\bibitem{conneau2016very}
A.~Conneau, H.~Schwenk, L.~Barrault, and Y.~Lecun, ``Very deep convolutional
  networks for text classification,'' \emph{arxiv:1606.01781}, 2016.

\bibitem{silver2016mastering}
D.~Silver, A.~Huang, C.~J. Maddison, A.~Guez, L.~Sifre, G.~Van Den~Driessche,
  J.~Schrittwieser, I.~Antonoglou, V.~Panneershelvam, M.~Lanctot \emph{et~al.},
  ``Mastering the game of go with deep neural networks and tree search,''
  \emph{nature}, vol. 529, no. 7587, 2016.

\bibitem{butler2018machine}
K.~T. Butler, D.~W. Davies, H.~Cartwright, O.~Isayev, and A.~Walsh, ``Machine
  learning for molecular and materials science,'' \emph{Nature}, vol. 559, no.
  7715, 2018.

\bibitem{ching2018opportunities}
T.~Ching, D.~S. Himmelstein, B.~K. Beaulieu-Jones, A.~A. Kalinin, B.~T. Do,
  G.~P. Way, E.~Ferrero, P.-M. Agapow, M.~Zietz, M.~M. Hoffman \emph{et~al.},
  ``Opportunities and obstacles for deep learning in biology and medicine,''
  \emph{J. Royal Society Interface}, vol.~15, no. 141, 2018.

\bibitem{kutz2017deep}
J.~N. Kutz, ``Deep learning in fluid dynamics,'' \emph{J. Fluid Mechanics},
  vol. 814, 2017.

\bibitem{brundage2020toward}
M.~Brundage, S.~Avin, J.~Wang, H.~Belfield, G.~Krueger, G.~Hadfield, H.~Khlaaf,
  J.~Yang, H.~Toner, R.~Fong \emph{et~al.}, ``Toward trustworthy ai
  development: mechanisms for supporting verifiable claims,''
  \emph{arXiv:2004.07213}, 2020.

\bibitem{roscher2019explainable}
R.~Roscher, B.~Bohn, M.~F. Duarte, and J.~Garcke, ``Explainable machine
  learning for scientific insights and discoveries,'' \emph{arxiv:1905.08883},
  2019.

\bibitem{diligenti2017integrating}
M.~Diligenti, S.~Roychowdhury, and M.~Gori, ``Integrating prior knowledge into
  deep learning,'' in \emph{Int. Conf. on Machine Learning and Applications
  (ICMLA)}.\hskip 1em plus 0.5em minus 0.4em\relax IEEE, 2017.

\bibitem{xu2017semantic}
J.~Xu, Z.~Zhang, T.~Friedman, Y.~Liang, and G.~V.~d. Broeck, ``A semantic loss
  function for deep learning with symbolic knowledge,''
  \emph{arxiv:1711.11157}, 2017.

\bibitem{karpatne2017physics}
A.~Karpatne, W.~Watkins, J.~Read, and V.~Kumar, ``Physics-guided neural
  networks (pgnn): An application in lake temperature modeling,''
  \emph{arxiv:1710.11431}, 2017.

\bibitem{stewart2017label}
R.~Stewart and S.~Ermon, ``Label-free supervision of neural networks with
  physics and domain knowledge,'' in \emph{Conf. Artificial
  Intelligence}.\hskip 1em plus 0.5em minus 0.4em\relax AAAI, 2017.

\bibitem{battaglia2016interaction}
P.~Battaglia, R.~Pascanu, M.~Lai, D.~J. Rezende \emph{et~al.}, ``Interaction
  networks for learning about objects, relations and physics,'' in \emph{Neural
  Information Processing Systems (NIPS)}, 2016.

\bibitem{marino2017more}
K.~Marino, R.~Salakhutdinov, and A.~Gupta, ``The more you know: Using knowledge
  graphs for image classification,'' in \emph{Conf. Computer Vision and Pattern
  Recognition (CVPR)}.\hskip 1em plus 0.5em minus 0.4em\relax IEEE, 2017.

\bibitem{jiang2018hybrid}
C.~Jiang, H.~Xu, X.~Liang, and L.~Lin, ``Hybrid knowledge routed modules for
  large-scale object detection,'' in \emph{Neural Information Processing
  Systems (NIPS)}, 2018.

\bibitem{cully2015robots}
A.~Cully, J.~Clune, D.~Tarapore, and J.-B. Mouret, ``Robots that can adapt like
  animals,'' \emph{Nature}, vol. 521, no. 7553, 2015.

\bibitem{lee2019spigan}
K.-H. Lee, J.~Li, A.~Gaidon, and G.~Ros, ``Spigan: Privileged adversarial
  learning from simulation,'' in \emph{Int. Conf. Learning Representations
  (ICLR)}, 2019.

\bibitem{pfrommer2018optimisation}
J.~Pfrommer, C.~Zimmerling, J.~Liu, L.~K{\"a}rger, F.~Henning, and J.~Beyerer,
  ``Optimisation of manufacturing process parameters using deep neural networks
  as surrogate models,'' \emph{Procedia CIRP}, vol.~72, no.~1, 2018.

\bibitem{raissi2017physics1}
M.~Raissi, P.~Perdikaris, and G.~E. Karniadakis, ``Physics informed deep
  learning (part i): Data-driven solutions of nonlinear partial differential
  equations,'' \emph{arxiv:1711.10561}, 2017.

\bibitem{diligenti2017semantic}
M.~Diligenti, M.~Gori, and C.~Sacca, ``Semantic-based regularization for
  learning and inference,'' \emph{Artificial Intelligence}, vol. 244, 2017.

\bibitem{karpatne2017theory}
A.~Karpatne, G.~Atluri, J.~H. Faghmous, M.~Steinbach, A.~Banerjee, A.~Ganguly,
  S.~Shekhar, N.~Samatova, and V.~Kumar, ``Theory-guided data science: A new
  paradigm for scientific discovery from data,'' \emph{Trans. Knowledge and
  Data Engineering}, vol.~29, no.~10, 2017.

\bibitem{lauer2008incorporating}
F.~Lauer and G.~Bloch, ``Incorporating prior knowledge in support vector
  machines for classification: A review,'' \emph{Neurocomputing}, vol.~71, no.
  7-9, 2008.

\bibitem{battaglia2018relational}
P.~W. Battaglia, J.~B. Hamrick, V.~Bapst, A.~Sanchez-Gonzalez, V.~Zambaldi,
  M.~Malinowski, A.~Tacchetti, D.~Raposo, A.~Santoro, R.~Faulkner
  \emph{et~al.}, ``Relational inductive biases, deep learning, and graph
  networks,'' \emph{arxiv:1806.01261}, 2018.

\bibitem{steup2018epistemology}
M.~Steup, ``Epistemologyp,'' in \emph{The Stanford Encyclopedia of Philosophy
  (Winter 2018 Edition), Edward N. Zalta (ed.)}, 2018.

\bibitem{zagzebski2017knowledge}
L.~Zagzebski, \emph{{What is Knowledge?}}\hskip 1em plus 0.5em minus
  0.4em\relax John Wiley {\&} Sons, 2017.

\bibitem{machamer2008blackwell}
P.~Machamer and M.~Silberstein, \emph{{The Blackwell guide to the philosophy of
  science}}.\hskip 1em plus 0.5em minus 0.4em\relax John Wiley {\&} Sons, 2008,
  vol.~19.

\bibitem{fayyad1996data}
U.~Fayyad, G.~Piatetsky-Shapiro, and P.~Smyth, ``From data mining to knowledge
  discovery in databases,'' \emph{AI magazine}, vol.~17, no.~3, 1996.

\bibitem{Kahneman2011}
D.~Kahneman, \emph{Thinking, Fast and Slow}.\hskip 1em plus 0.5em minus
  0.4em\relax Macmillan, 2011.

\bibitem{lake2017building}
B.~M. Lake, T.~D. Ullman, J.~B. Tenenbaum, and S.~J. Gershman, ``{Building
  machines that learn and think like people},'' \emph{Behavioral and brain
  sciences}, vol.~40, 2017.

\bibitem{gauch2003scientific}
H.~G. Gauch, \emph{{Scientific method in practice}}.\hskip 1em plus 0.5em minus
  0.4em\relax Cambridge University Press, 2003.

\bibitem{abu2012learning}
Y.~S. Abu-Mostafa, M.~Magdon-Ismail, and H.-T. Lin, \emph{Learning from
  data}.\hskip 1em plus 0.5em minus 0.4em\relax AMLBook, 2012, vol.~4.

\bibitem{muralidhar2018incorporating}
N.~Muralidhar, M.~R. Islam, M.~Marwah, A.~Karpatne, and N.~Ramakrishnan,
  ``Incorporating prior domain knowledge into deep neural networks,'' in
  \emph{Int. Conf. Big Data}.\hskip 1em plus 0.5em minus 0.4em\relax IEEE,
  2018.

\bibitem{lu2017physics}
Y.~Lu, M.~Rajora, P.~Zou, and S.~Liang, ``Physics-embedded machine learning:
  Case study with electrochemical micro-machining,'' \emph{Machines}, vol.~5,
  no.~1, 2017.

\bibitem{heese2019good}
R.~Heese, M.~Walczak, L.~Morand, D.~Helm, and M.~Bortz, ``The good, the bad and
  the ugly: Augmenting a black-box model with expert knowledge,'' in \emph{Int.
  Conf. Artificial Neural Networks (ICANN)}.\hskip 1em plus 0.5em minus
  0.4em\relax Springer, 2019.

\bibitem{fung2003knowledge}
G.~M. Fung, O.~L. Mangasarian, and J.~W. Shavlik, ``Knowledge-based support
  vector machine classifiers,'' in \emph{Neural Information Processing Systems
  (NIPS)}, 2003.

\bibitem{mangasarian2008nonlinear}
O.~L. Mangasarian and E.~W. Wild, ``Nonlinear knowledge-based classification,''
  \emph{Trans. Neural Networks}, vol.~19, no.~10, 2008.

\bibitem{ramamurthy2019leveraging}
R.~Ramamurthy, C.~Bauckhage, R.~Sifa, J.~Sch{\"u}cker, and S.~Wrobel,
  ``Leveraging domain knowledge for reinforcement learning using mmc
  architectures,'' in \emph{Int. Conf. Artificial Neural Networks
  (ICANN)}.\hskip 1em plus 0.5em minus 0.4em\relax Springer, 2019.

\bibitem{vonkurnatowski2020compensating}
M.~von Kurnatowski, J.~Schmid, P.~Link, R.~Zache, L.~Morand, T.~Kraft,
  I.~Schmidt, and A.~Stoll, ``Compensating data shortages in manufacturing with
  monotonicity knowledge,'' \emph{arXiv:2010.15955}, 2020.

\bibitem{bauckhage2018informed}
C.~Bauckhage, C.~Ojeda, J.~Sch{\"u}cker, R.~Sifa, and S.~Wrobel, ``Informed
  machine learning through functional composition.''

\bibitem{king2018deep}
R.~King, O.~Hennigh, A.~Mohan, and M.~Chertkov, ``From deep to physics-informed
  learning of turbulence: Diagnostics,'' \emph{arxiv:1810.07785}, 2018.

\bibitem{jeong2015data}
S.~Jeong, B.~Solenthaler, M.~Pollefeys, M.~Gross \emph{et~al.}, ``Data-driven
  fluid simulations using regression forests,'' \emph{ACM Trans. Graphics},
  vol.~34, no.~6, 2015.

\bibitem{yang2018physics}
Y.~Yang and P.~Perdikaris, ``Physics-informed deep generative models,''
  \emph{arxiv:1812.03511}, 2018.

\bibitem{debezenac}
E.~de~Bezenac, A.~Pajot, and P.~Gallinari, ``Deep learning for physical
  processes: Incorporating prior scientific knowledge,''
  \emph{arxiv:1711.07970}, 2017.

\bibitem{lagaris1998artificial}
I.~E. Lagaris, A.~Likas, and D.~I. Fotiadis, ``Artificial neural networks for
  solving ordinary and partial differential equations,'' \emph{Trans. Neural
  Networks}, vol.~9, no.~5, 1998.

\bibitem{zhu2019}
Y.~Zhu, N.~Zabaras, P.-S. Koutsourelakis, and P.~Perdikaris,
  ``Physics-constrained deep learning for high-dimensional surrogate modeling
  and uncertainty quantification without labeled data,'' \emph{J. Computational
  Physics}, vol. 394, 2019.

\bibitem{psichogios1992hybrid}
D.~C. Psichogios and L.~H. Ungar, ``A hybrid neural network-first principles
  approach to process modeling,'' \emph{AIChE Journal}, vol.~38, no.~10, pp.
  1499--1511, 1992.

\bibitem{lutter2019deep}
M.~Lutter, C.~Ritter, and J.~Peters, ``Deep lagrangian networks: Using physics
  as model prior for deep learning,'' \emph{arxiv:1907.04490}, 2019.

\bibitem{belbute-peres2018}
F.~D.~A. Belbute-peres, K.~R. Allen, K.~A. Smith, and J.~B. Tenenbaum,
  ``End-to-end differentiable physics for learning and control,'' in
  \emph{Neural Information Processing Systems (NIPS)}, 2018.

\bibitem{ling2016reynolds}
J.~Ling, A.~Kurzawski, and J.~Templeton, ``Reynolds averaged turbulence
  modelling using deep neural networks with embedded invariance,'' \emph{J.
  Fluid Mechanics}, vol. 807, 2016.

\bibitem{butter2018deep}
A.~Butter, G.~Kasieczka, T.~Plehn, and M.~Russell, ``Deep-learned top tagging
  with a lorentz layer,'' \emph{SciPost Phys}, vol.~5, no.~28, 2018.

\bibitem{wu2018physics}
J.-L. Wu, H.~Xiao, and E.~Paterson, ``Physics-informed machine learning
  approach for augmenting turbulence models: A comprehensive framework,''
  \emph{Physical Review Fluids}, vol.~3, no.~7, 2018.

\bibitem{towell1994knowledge}
G.~G. Towell and J.~W. Shavlik, ``Knowledge-based artificial neural networks,''
  \emph{Artificial Intelligence}, vol.~70, no. 1-2, 1994.

\bibitem{garcez1999connectionist}
A.~S.~d. Garcez and G.~Zaverucha, ``The connectionist inductive learning and
  logic programming system,'' \emph{Applied Intelligence}, vol.~11, no.~1,
  1999.

\bibitem{ma2018multi}
T.~Ma and A.~Zhang, ``Multi-view factorization autoencoder with network
  constraints for multi-omic integrative analysis,'' in \emph{Int. Conf.
  Bioinformatics and Biomedicine (BIBM)}.\hskip 1em plus 0.5em minus
  0.4em\relax IEEE, 2018.

\bibitem{che2015deep}
Z.~Che, D.~Kale, W.~Li, M.~T. Bahadori, and Y.~Liu, ``Deep computational
  phenotyping,'' in \emph{Int. Conf. Knowledge Discovery and Data Mining
  (SIGKDD)}.\hskip 1em plus 0.5em minus 0.4em\relax ACM, 2015.

\bibitem{messaoud2009integrating}
M.~B. Messaoud, P.~Leray, and N.~B. Amor, ``Integrating ontological knowledge
  for iterative causal discovery and visualization,'' in \emph{European Conf.
  Symbolic and Quantitative Approaches to Reasoning and Uncertainty}.\hskip 1em
  plus 0.5em minus 0.4em\relax Springer, 2009.

\bibitem{borboudakis2012incorporating}
G.~Borboudakis and I.~Tsamardinos, ``Incorporating causal prior knowledge as
  path-constraints in bayesian networks and maximal ancestral graphs,''
  \emph{arxiv:1206.6390}, 2012.

\bibitem{deist2019simulation}
T.~Deist, A.~Patti, Z.~Wang, D.~Krane, T.~Sorenson, and D.~Craft, ``Simulation
  assisted machine learning.'' \emph{Bioinformatics (Oxford, England)}, 2019.

\bibitem{kim2007hybrid}
H.~S. Kim, M.~Koc, and J.~Ni, ``A hybrid multi-fidelity approach to the optimal
  design of warm forming processes using a knowledge-based artificial neural
  network,'' \emph{Int. J. Machine Tools and Manufacture}, vol.~47, no.~2,
  2007.

\bibitem{hautier2010finding}
G.~Hautier, C.~C. Fischer, A.~Jain, T.~Mueller, and G.~Ceder, ``Finding
  nature’s missing ternary oxide compounds using machine learning and density
  functional theory,'' \emph{Chemistry of Materials}, vol.~22, no.~12, 2010.

\bibitem{chang2016compositional}
M.~B. Chang, T.~Ullman, A.~Torralba, and J.~B. Tenenbaum, ``A compositional
  object-based approach to learning physical dynamics,''
  \emph{arxiv:1612.00341}, 2016.

\bibitem{choi2017gram}
E.~Choi, M.~T. Bahadori, L.~Song, W.~F. Stewart, and J.~Sun, ``Gram:
  Graph-based attention model for healthcare representation learning,'' in
  \emph{Int. Conf. Knowledge Discovery and Data Mining (SIGKDD)}.\hskip 1em
  plus 0.5em minus 0.4em\relax ACM, 2017.

\bibitem{lerer2016learning}
A.~Lerer, S.~Gross, and R.~Fergus, ``Learning physical intuition of block
  towers by example,'' \emph{arxiv:1603.01312}, 2016.

\bibitem{rai2019using}
A.~Rai, R.~Antonova, F.~Meier, and C.~G. Atkeson, ``Using simulation to improve
  sample-efficiency of bayesian optimization for bipedal robots.'' \emph{J.
  Machine Learning Research}, vol.~20, no.~49, 2019.

\bibitem{du2018learning}
Y.~Du, Z.~Liu, H.~Basevi, A.~Leonardis, B.~Freeman, J.~Tenenbaum, and J.~Wu,
  ``Learning to exploit stability for 3d scene parsing,'' in \emph{Neural
  Information Processing Systems (NIPS)}, 2018.

\bibitem{shrivastava2017learning}
A.~Shrivastava, T.~Pfister, O.~Tuzel, J.~Susskind, W.~Wang, and R.~Webb,
  ``Learning from simulated and unsupervised images through adversarial
  training,'' in \emph{Conf. Computer Vision and Pattern Recognition
  (CVPR)}.\hskip 1em plus 0.5em minus 0.4em\relax IEEE, 2017.

\bibitem{wang1997knowledge}
F.~Wang and Q.-J. Zhang, ``Knowledge-based neural models for microwave
  design,'' \emph{Trans. Microwave Theory and Techniques}, vol.~45, no.~12,
  1997.

\bibitem{leary2003knowledge}
S.~J. Leary, A.~Bhaskar, and A.~J. Keane, ``A knowledge-based approach to
  response surface modelling in multifidelity optimization,'' \emph{J. Global
  Optimization}, vol.~26, no.~3, 2003.

\bibitem{cohen2016group}
T.~S. Cohen and M.~Welling, ``Group equivariant convolutional networks,'' in
  \emph{Int. Conf. Machine Learning (ICML)}, 2016.

\bibitem{dieleman2016exploiting}
S.~Dieleman, J.~De~Fauw, and K.~Kavukcuoglu, ``Exploiting cyclic symmetry in
  convolutional neural networks,'' \emph{arxiv:1602.02660}, 2016.

\bibitem{scholkopf1998prior}
B.~Sch{\"o}lkopf, P.~Simard, A.~J. Smola, and V.~Vapnik, ``Prior knowledge in
  support vector kernels,'' in \emph{Neural Information Processing Systems
  (NIPS)}, 1998.

\bibitem{yet2014combining}
B.~Yet, Z.~B. Perkins, T.~E. Rasmussen, N.~R. Tai, and D.~W.~R. Marsh,
  ``Combining data and meta-analysis to build bayesian networks for clinical
  decision support,'' \emph{J. Biomedical Informatics}, vol.~52, 2014.

\bibitem{li2018deep}
J.~Li, Z.~Yang, H.~Liu, and D.~Cai, ``Deep rotation equivariant network,''
  \emph{Neurocomputing}, vol. 290, 2018.

\bibitem{worrall2017harmonic}
D.~E. Worrall, S.~J. Garbin, D.~Turmukhambetov, and G.~J. Brostow, ``Harmonic
  networks: Deep translation and rotation equivariance,'' in \emph{Conf.
  Computer Vision and Pattern Recognition (CVPR)}.\hskip 1em plus 0.5em minus
  0.4em\relax IEEE, 2017.

\bibitem{niyogi1998incorporating}
P.~Niyogi, F.~Girosi, and T.~Poggio, ``Incorporating prior information in
  machine learning by creating virtual examples,'' \emph{Proc. of the IEEE},
  vol.~86, no.~11, 1998.

\bibitem{schiegg2012markov}
M.~Schiegg, M.~Neumann, and K.~Kersting, ``Markov logic mixtures of gaussian
  processes: Towards machines reading regression data,'' in \emph{Artificial
  Intelligence and Statistics}, 2012.

\bibitem{sachan2018learning}
M.~Sachan, K.~A. Dubey, T.~M. Mitchell, D.~Roth, and E.~P. Xing, ``Learning
  pipelines with limited data and domain knowledge: A study in parsing physics
  problems,'' in \emph{Neural Information Processing Systems (NIPS)}, 2018.

\bibitem{zhou2018commonsense}
H.~Zhou, T.~Young, M.~Huang, H.~Zhao, J.~Xu, and X.~Zhu, ``Commonsense
  knowledge aware conversation generation with graph attention.'' in \emph{Int.
  Joint Conf. Artificial Intelligence (IJCAI)}, 2018.

\bibitem{mintz2009distant}
M.~Mintz, S.~Bills, R.~Snow, and D.~Jurafsky, ``Distant supervision for
  relation extraction without labeled data,'' in \emph{Association for
  Computational Linguistics (ACL)}, 2009.

\bibitem{liang2018symbolic}
X.~Liang, Z.~Hu, H.~Zhang, L.~Lin, and E.~P. Xing, ``Symbolic graph reasoning
  meets convolutions,'' in \emph{Neural Information Processing Systems (NIPS)},
  2018.

\bibitem{bergman2019symmetry}
D.~L. Bergman, ``Symmetry constrained machine learning,'' in \emph{SAI
  Intelligent Systems Conf.}\hskip 1em plus 0.5em minus 0.4em\relax Springer,
  2019.

\bibitem{chang2007guiding}
M.-W. Chang, L.~Ratinov, and D.~Roth, ``Guiding semi-supervision with
  constraint-driven learning,'' in \emph{Association for Computational
  Linguistics (ACL)}, 2007.

\bibitem{hu2016deep}
Z.~Hu, Z.~Yang, R.~Salakhutdinov, and E.~Xing, ``Deep neural networks with
  massive learned knowledge,'' in \emph{Conf. Empirical Methods in Natural
  Language Processing}, 2016.

\bibitem{hu2016harnessing}
Z.~Hu, X.~Ma, Z.~Liu, E.~Hovy, and E.~Xing, ``Harnessing deep neural networks
  with logic rules,'' \emph{arxiv:1603.06318}, 2016.

\bibitem{zhang2019ernie}
Z.~Zhang, X.~Han, Z.~Liu, X.~Jiang, M.~Sun, and Q.~Liu, ``Ernie: Enhanced
  language representation with informative entities,'' \emph{arxiv:1905.07129},
  2019.

\bibitem{mrkvsic2016counter}
N.~Mrk{\v{s}}i{\'c}, D.~O. S{\'e}aghdha, B.~Thomson, M.~Ga{\v{s}}i{\'c},
  L.~Rojas-Barahona, P.-H. Su, D.~Vandyke, T.-H. Wen, and S.~Young,
  ``Counter-fitting word vectors to linguistic constraints,''
  \emph{arxiv:1603.00892}, 2016.

\bibitem{bian2014knowledge}
J.~Bian, B.~Gao, and T.-Y. Liu, ``Knowledge-powered deep learning for word
  embedding,'' in \emph{Joint European Conf. machine learning and knowledge
  discovery in databases}.\hskip 1em plus 0.5em minus 0.4em\relax Springer,
  2014.

\bibitem{francca2014fast}
M.~V. Fran{\c{c}}a, G.~Zaverucha, and A.~S.~d. Garcez, ``Fast relational
  learning using bottom clause propositionalization with artificial neural
  networks,'' \emph{Machine Learning}, vol.~94, no.~1, 2014.

\bibitem{richardson2006markov}
M.~Richardson and P.~Domingos, ``Markov logic networks,'' \emph{Machine
  Learning}, vol.~62, no. 1-2, 2006.

\bibitem{kimmig2012short}
A.~Kimmig, S.~Bach, M.~Broecheler, B.~Huang, and L.~Getoor, ``A short
  introduction to probabilistic soft logic,'' in \emph{NIPS Workshop on
  Probabilistic Programming: Foundations and Applications}, 2012.

\bibitem{glavavs2018explicit}
G.~Glava{\v{s}} and I.~Vuli{\'c}, ``Explicit retrofitting of distributional
  word vectors,'' in \emph{Association for Computational Linguistics (ACL)},
  2018.

\bibitem{fang2017object}
Y.~Fang, K.~Kuan, J.~Lin, C.~Tan, and V.~Chandrasekhar, ``Object detection
  meets knowledge graphs,'' 2017.

\bibitem{peters2019knowledge}
M.~E. Peters, M.~Neumann, R.~Logan, R.~Schwartz, V.~Joshi, S.~Singh, and N.~A.
  Smith, ``Knowledge enhanced contextual word representations,'' in \emph{Conf.
  Empirical Methods in Natural Language Processing (EMNLP), Int. Joint Conf.
  Natural Language Processing (IJCNLP)}, 2019.

\bibitem{Campos2007bayesian}
L.~M. de~Campos and J.~G. Castellano, ``Bayesian network learning algorithms
  using structural restrictions,'' \emph{Int. J. Approximate Reasoning},
  vol.~45, no.~2, 2007.

\bibitem{constantinou2016integrating}
A.~C. Constantinou, N.~Fenton, and M.~Neil, ``Integrating expert knowledge with
  data in bayesian networks: Preserving data-driven expectations when the
  expert variables remain unobserved,'' \emph{Expert Systems with
  Applications}, vol.~56, 2016.

\bibitem{choo2013utopian}
J.~Choo, C.~Lee, C.~K. Reddy, and H.~Park, ``Utopian: User-driven topic
  modeling based on interactive nonnegative matrix factorization,''
  \emph{Trans. Visualization and Computer Graphics}, vol.~19, no.~12, 2013.

\bibitem{knox2009interactively}
W.~B. Knox and P.~Stone, ``Interactively shaping agents via human
  reinforcement: The tamer framework,'' in \emph{Int. Conf. Knowledge Cature
  (K-CAP)}.\hskip 1em plus 0.5em minus 0.4em\relax ACM, 2009.

\bibitem{kaplan2017beating}
R.~Kaplan, C.~Sauer, and A.~Sosa, ``Beating atari with natural language guided
  reinforcement learning,'' \emph{arxiv:1704.05539}, 2017.

\bibitem{heckerman1995learning}
D.~Heckerman, D.~Geiger, and D.~M. Chickering, ``Learning bayesian networks:
  The combination of knowledge and statistical data,'' \emph{Machine Learning},
  vol.~20, no.~3, 1995.

\bibitem{richardson2003learning}
M.~Richardson and P.~Domingos, ``Learning with knowledge from multiple
  experts,'' in \emph{Int. Conf. Machine Learning (ICML)}, 2003.

\bibitem{feelders2006learning}
A.~Feelders and L.~C. Van~der Gaag, ``Learning bayesian network parameters
  under order constraints,'' \emph{Int. J. Approximate Reasoning}, vol.~42, no.
  1-2, 2006.

\bibitem{christiano2017deep}
P.~F. Christiano, J.~Leike, T.~Brown, M.~Martic, S.~Legg, and D.~Amodei, ``Deep
  reinforcement learning from human preferences,'' in \emph{Neural Information
  Processing Systems (NIPS)}, 2017.

\bibitem{hester2018deep}
T.~Hester, M.~Vecerik, O.~Pietquin, M.~Lanctot, T.~Schaul, B.~Piot, D.~Horgan,
  J.~Quan, A.~Sendonaris, I.~Osband \emph{et~al.}, ``Deep q-learning from
  demonstrations,'' in \emph{Conf. Artificial Intelligence}.\hskip 1em plus
  0.5em minus 0.4em\relax AAAI, 2018.

\bibitem{brown2012dis}
E.~T. Brown, J.~Liu, C.~E. Brodley, and R.~Chang, ``Dis-function: Learning
  distance functions interactively,'' in \emph{Conf. Visual Analytics Science
  and Technology (VAST)}.\hskip 1em plus 0.5em minus 0.4em\relax IEEE, 2012.

\bibitem{yet2014not}
B.~Yet, Z.~Perkins, N.~Fenton, N.~Tai, and W.~Marsh, ``Not just data: A method
  for improving prediction with knowledge,'' \emph{J. Biomedical Informatics},
  vol.~48, 2014.

\bibitem{fails2003interactive}
J.~A. Fails and D.~R. Olsen~Jr, ``Interactive machine learning,'' in \emph{Int.
  Conf. Intelligent User Interfaces}.\hskip 1em plus 0.5em minus 0.4em\relax
  ACM, 2003.

\bibitem{rieger2019interpretations}
L.~Rieger, C.~Singh, W.~J. Murdoch, and B.~Yu, ``Interpretations are useful:
  penalizing explanations to align neural networks with prior knowledge,''
  \emph{arXiv:1909.13584}, 2019.

\bibitem{schramowski2020right}
P.~Schramowski, W.~Stammer, S.~Teso, A.~Brugger, H.-G. Luigs, A.-K. Mahlein,
  and K.~Kersting, ``Right for the wrong scientific reasons: Revising deep
  networks by interacting with their explanations,'' \emph{arXiv:2001.05371},
  2020.

\bibitem{vonrueden2020streetmap}
L.~von Rueden, T.~Wirtz, F.~Hueger, J.~D. Schneider, N.~Piatkowski, and
  C.~Bauckhage, ``Street-map based validation of semantic segmentation in
  autonomous driving,'' in \emph{Int. Conf. Pattern Recognition (ICPR)}.\hskip
  1em plus 0.5em minus 0.4em\relax IEEE, 2020.

\bibitem{bronstein2017geometric}
M.~M. Bronstein, J.~Bruna, Y.~LeCun, A.~Szlam, and P.~Vandergheynst,
  ``Geometric deep learning: Going beyond euclidean data,'' \emph{Signal
  Processing}, vol.~34, no.~4, 2017.

\bibitem{chang2012structured}
M.-W. Chang, L.~Ratinov, and D.~Roth, ``Structured learning with constrained
  conditional models,'' \emph{Machine Learning}, vol.~88, no.~3, 2012.

\bibitem{sridhar2015joint}
D.~Sridhar, J.~Foulds, B.~Huang, L.~Getoor, and M.~Walker, ``Joint models of
  disagreement and stance in online debate,'' in \emph{Association for
  Computational Linguistics (ACL)}, 2015.

\bibitem{garcez2019neural}
A.~S.~d. Garcez, M.~Gori, L.~C. Lamb, L.~Serafini, M.~Spranger, and S.~N. Tran,
  ``Neural-symbolic computing: An effective methodology for principled
  integration of machine learning and reasoning,'' \emph{arxiv:1905.06088},
  2019.

\bibitem{Raedt2016Statistical}
L.~D. Raedt, K.~Kersting, and S.~Natarajan, \emph{Statistical Relational
  Artificial Intelligence: Logic, Probability, and Computation}.\hskip 1em plus
  0.5em minus 0.4em\relax Morgan \& Claypool Publishers, 2016.

\bibitem{speer2013conceptnet}
R.~Speer and C.~Havasi, ``Conceptnet 5: A large semantic network for relational
  knowledge,'' in \emph{The People’s Web Meets NLP}.\hskip 1em plus 0.5em
  minus 0.4em\relax Springer, 2013, pp. 161--176.

\bibitem{miller1995wordnet}
G.~A. Miller, ``Wordnet: A lexical database for english,'' \emph{Communications
  ACM}, vol.~38, no.~11, 1995.

\bibitem{mitchell2018never}
T.~Mitchell, W.~Cohen, E.~Hruschka, P.~Talukdar, B.~Yang, J.~Betteridge,
  A.~Carlson, B.~Dalvi, M.~Gardner, B.~Kisiel \emph{et~al.}, ``Never-ending
  learning,'' \emph{Communications ACM}, vol.~61, no.~5, 2018.

\bibitem{devlin2018bert}
J.~Devlin, M.-W. Chang, K.~Lee, and K.~Toutanova, ``Bert: Pre-training of deep
  bidirectional transformers for language understanding,''
  \emph{arxiv:1810.04805}, 2018.

\bibitem{ye2019distant}
Z.-X. Ye and Z.-H. Ling, ``Distant supervision relation extraction with
  intra-bag and inter-bag attentions,'' \emph{arxiv:1904.00143}, 2019.

\bibitem{mikolov2013efficient}
T.~Mikolov, K.~Chen, G.~Corrado, and J.~Dean, ``Efficient estimation of word
  representations in vector space,'' \emph{arxiv:1301.3781}, 2013.

\bibitem{massa2010gene}
M.~S. Massa, M.~Chiogna, and C.~Romualdi, ``Gene set analysis exploiting the
  topology of a pathway,'' \emph{BMC systems biology}, vol.~4, no.~1, 2010.

\bibitem{angelopoulos2008bayesian}
N.~Angelopoulos and J.~Cussens, ``Bayesian learning of bayesian networks with
  informative priors,'' \emph{Annals of Mathematics and Artificial
  Intelligence}, vol.~54, no. 1-3, 2008.

\bibitem{piatkowski2013spatio}
N.~Piatkowski, S.~Lee, and K.~Morik, ``Spatio-temporal random fields:
  Compressible representation and distributed estimation,'' \emph{Machine
  Learning}, vol.~93, no.~1, 2013.

\bibitem{fischer20no}
R.~Fischer, N.~Piatkowski, C.~Pelletier, G.~I. Webb, F.~Petitjean, and
  K.~Morik, ``No cloud on the horizon: Probabilistic gap filling in satellite
  image series,'' in \emph{Int. Conf. Data Science and Advanced Analytics
  (DSAA)}, 2020.

\bibitem{settles2009active}
B.~Settles, ``Active learning literature survey,'' University of
  Wisconsin--Madison, Computer Sciences Technical Report 1648, 2009.

\bibitem{keim2008visual}
D.~Keim, G.~Andrienko, J.-D. Fekete, C.~G{\"o}rg, J.~Kohlhammer, and
  G.~Melan{\c{c}}on, ``Visual analytics: Definition, process, and challenges,''
  in \emph{Information visualization}.\hskip 1em plus 0.5em minus 0.4em\relax
  Springer, 2008.

\bibitem{minsky1991logical}
M.~L. Minsky, ``Logical versus analogical or symbolic versus connectionist or
  neat versus scruffy,'' \emph{AI magazine}, vol.~12, no.~2, 1991.

\bibitem{kalnay2003atmospheric}
E.~Kalnay, \emph{Atmospheric modeling, data assimilation and
  predictability}.\hskip 1em plus 0.5em minus 0.4em\relax Cambridge university
  press, 2003.

\bibitem{reich2015probabilistic}
S.~Reich and C.~Cotter, \emph{Probabilistic forecasting and Bayesian data
  assimilation}.\hskip 1em plus 0.5em minus 0.4em\relax Cambridge University
  Press, 2015.

\bibitem{janner2017self}
M.~Janner, J.~Wu, T.~D. Kulkarni, I.~Yildirim, and J.~B. Tenenbaum,
  ``Self-supervised intrinsic image decomposition,'' in \emph{Neural
  Information Processing Systems (NIPS)}, 2017.

\bibitem{wang2020generalizing}
Y.~Wang, Q.~Yao, J.~T. Kwok, and L.~M. Ni, ``Generalizing from a few examples:
  A survey on few-shot learning,'' \emph{ACM Computing Surveys (CSUR)},
  vol.~53, no.~3, 2020.

\bibitem{cucker2007learning}
F.~Cucker and D.~X. Zhou, \emph{Learning theory: an approximation theory
  viewpoint}.\hskip 1em plus 0.5em minus 0.4em\relax Cambridge University
  Press, 2007, vol.~24.

\bibitem{steinwart2008support}
I.~Steinwart and A.~Christmann, \emph{Support vector machines}.\hskip 1em plus
  0.5em minus 0.4em\relax Springer Science \& Business Media, 2008.

\bibitem{cucker2002best}
F.~Cucker and S.~Smale, ``Best choices for regularization parameters in
  learning theory: on the bias-variance problem,'' \emph{Foundations of
  computational Mathematics}, vol.~2, no.~4, 2002.

\bibitem{lapidus2011numerical}
L.~Lapidus and G.~F. Pinder, \emph{Numerical solution of partial differential
  equations in science and engineering}.\hskip 1em plus 0.5em minus 0.4em\relax
  John Wiley \& Sons, 2011.

\bibitem{wulfmeier2017addressing}
M.~Wulfmeier, A.~Bewley, and I.~Posner, ``Addressing appearance change in
  outdoor robotics with adversarial domain adaptation,'' in \emph{Int. Conf.
  Intelligent Robots and Systems (IROS)}.\hskip 1em plus 0.5em minus
  0.4em\relax IEEE, 2017.

\bibitem{peng2018sim}
X.~B. Peng, M.~Andrychowicz, W.~Zaremba, and P.~Abbeel, ``Sim-to-real transfer
  of robotic control with dynamics randomization,'' in \emph{Int. Conf.
  Robotics and Automation (ICRA)}.\hskip 1em plus 0.5em minus 0.4em\relax IEEE,
  2018.

\bibitem{vonrueden2020combining}
L.~von Rueden, S.~Mayer, R.~Sifa, C.~Bauckhage, and J.~Garcke, ``Combining
  machine learning and simulation to a hybrid modelling approach: Current and
  future directions,'' in \emph{Int. Symp. Intelligent Data Analysis
  (IDA)}.\hskip 1em plus 0.5em minus 0.4em\relax Springer, 2020.

\bibitem{mcgarry1999hybrid}
K.~McGarry, S.~Wermter, and J.~MacIntyre, ``Hybrid neural systems: From simple
  coupling to fully integrated neural networks,'' \emph{Neural Computing
  Surveys}, vol.~2, no.~1, 1999.

\bibitem{sun2006connectionist}
R.~Sun, ``Connectionist implementationalism and hybrid systems,''
  \emph{Encyclopedia of Cognitive Science}, 2006.

\bibitem{garcez2020neurosymbolic}
A.~S.~d. Garcez and L.~C. Lamb, ``Neurosymbolic ai: The 3rd wave,''
  \emph{arXiv:2012.05876}, 2020.

\bibitem{dong2018imposing}
T.~Dong, C.~Bauckhage, H.~Jin, J.~Li, O.~Cremers, D.~Speicher, A.~B. Cremers,
  and J.~Zimmermann, ``Imposing category trees onto word-embeddings using a
  geometric construction,'' in \emph{Int. Conf. Learning Representations
  (ICLR)}, 2018.

\bibitem{bach2015Hinge}
S.~H. Bach, M.~Broecheler, B.~Huang, and L.~Getoor, ``{Hinge-loss markov random
  fields and probabilistic soft logic},'' \emph{arxiv:1505.04406}, 2015.

\bibitem{Embar2018ScalableSL}
V.~Embar, D.~Sridhar, G.~Farnadi, and L.~Getoor, ``Scalable structure learning
  for probabilistic soft logic,'' \emph{arXiv:1807.00973}.

\bibitem{blei2017variational}
D.~M. Blei, A.~Kucukelbir, and J.~D. McAuliffe, ``Variational inference: A
  review for statisticians,'' \emph{J. American Statistical Association}, vol.
  112, no. 518, 2017.

\bibitem{kingma2019introduction}
D.~P. Kingma and M.~Welling, ``An introduction to variational autoencoders,''
  \emph{arXiv:1906.02691}, 2019.

\bibitem{pearl2009causality}
J.~Pearl, \emph{Causality}.\hskip 1em plus 0.5em minus 0.4em\relax Cambridge
  university press, 2009.

\bibitem{kreutzer2020learning}
J.~Kreutzer, S.~Riezler, and C.~Lawrence, ``Learning from human feedback:
  Challenges for real-world reinforcement learning in nlp,''
  \emph{arXiv:2011.02511}, 2020.

\bibitem{dulac2019challenges}
G.~Dulac-Arnold, D.~Mankowitz, and T.~Hester, ``Challenges of real-world
  reinforcement learning,'' \emph{arXiv:1904.12901}, 2019.

\bibitem{kreutzer2018reliability}
J.~Kreutzer, J.~Uyheng, and S.~Riezler, ``Reliability and learnability of human
  bandit feedback for sequence-to-sequence reinforcement learning,'' in
  \emph{Association for Computational Linguistics (ACL)}, 2018.

\bibitem{gao2018april}
Y.~Gao, C.~M. Meyer, and I.~Gurevych, ``April: Interactively learning to
  summarise by combining active preference learning and reinforcement
  learning,'' in \emph{Conf. Empirical Methods in Natural Language Processing
  (EMNLP)}, 2018.

\bibitem{doshi2017towards}
F.~Doshi-Velez and B.~Kim, ``Towards a rigorous science of interpretable
  machine learning,'' \emph{arXiv:1702.08608}, 2017.

\end{thebibliography}
